%% file: main.tex
\documentclass{article}

\usepackage{microtype}
\usepackage{graphicx}
\usepackage{caption}
\usepackage{subcaption}
\usepackage[table]{xcolor}
\usepackage{booktabs} 
\usepackage{graphicx}
\usepackage{multirow}

\usepackage{hyperref}
\let\origaddcontentsline\addcontentsline

\usepackage[accepted]{icml2026}
\usepackage{amsmath}
\usepackage{amssymb}
\usepackage{mathtools}
\usepackage{amsthm}
\usepackage{enumitem}
\usepackage{multirow}
\usepackage{pifont}
\usepackage{soul}
\usepackage{hyperref}
\usepackage[capitalize,noabbrev]{cleveref}

\usepackage{tcolorbox}
\usepackage[toc,page,header]{appendix}
\usepackage{minitoc}
\usepackage{titletoc}

\theoremstyle{plain}
\newtheorem{theorem}{Theorem}[section]

\theoremstyle{definition}
\newtheorem{definition}[theorem]{Definition}

\theoremstyle{remark}

\usepackage[textsize=tiny]{todonotes}

\icmltitlerunning{\name{}}

\definecolor{purple}{HTML}{7161ef}

\definecolor{skylightblue}{HTML}{8ECAE6}
\definecolor{bluegreen}{HTML}{219EBC}
\definecolor{cerulean}{HTML}{126782}
\definecolor{dblue}{HTML}{023047}
\definecolor{amberflame}{HTML}{FFB703}
\definecolor{amberglow}{HTML}{FD9E02}
\definecolor{tigerorange}{HTML}{FB8500}

\hypersetup{
    colorlinks=true,
    linkcolor=dblue,   
    citecolor=dblue,   
    urlcolor=dblue,    
}

\colorlet{hlskybluecolor}{skylightblue!20}
\colorlet{hlbluegreencolor}{bluegreen!20}
\colorlet{hlambercolor}{amberflame!20}
\colorlet{hltigerorangecolor}{tigerorange!30}
\newcommand{\hlskyblue}[1]{{\sethlcolor{hlskybluecolor}\hl{\mbox{#1}}}}
\newcommand{\hlcerulean}[1]{{\sethlcolor{hlbluegreencolor}\hl{\mbox{#1}}}}
\newcommand{\hlamber}[1]{{\sethlcolor{hlambercolor}\hl{\mbox{#1}}}}

\newcommand\name[0]{\textbf{{\texttt{\textsc{IdEst}}}}}

\newcommand\grayout[1]{\textcolor{gray}{#1}}
\newcommand\best[1]{\textbf{\textcolor{black}{#1}}}

\newcommand\costst[1]{\ensuremath{L\left(#1\right)}}
\newcommand\costmst[1]{\ensuremath{\costst{\mstx{#1}}}}
\newcommand\mst[0]{\ensuremath{\mathrm{MST}}}
\newcommand\mstx[1]{\ensuremath{\mst{}{\left(#1\right)}}}

\newcommand\dimmst[0]{\ensuremath{\mathrm{dim}_\text{MST}}}
\newcommand\lu[0]{\ensuremath{\mathcal{L}_u}}
\newcommand\wasser[0]{\ensuremath{\mathcal{W}_2}}

\makeatletter
\newcommand{\hypbox}[2][]{%
  \begin{tcolorbox}[
      colframe=dblue!70,
      colback=white,
      boxrule=0.9pt,
      left=5pt,
      arc=2mm,
      right=5pt,top=2pt,bottom=2pt,
  ]
  \@ifnotempty{#1}{%
  \textcolor{dblue!90}{\textbf{#1}%
  #2}}
  \end{tcolorbox}%
}
\makeatother

\begin{document}

\twocolumn[
  \icmltitle{\name{}: Assessing Self-Supervised Learning Representations via Intrinsic Dimension}


  \icmlsetsymbol{equal}{*}

  \begin{icmlauthorlist}
    \icmlauthor{Julie Mordacq}{inria,x}
    \icmlauthor{Vicky Kalogeiton}{x}
    \icmlauthor{Steve Oudot}{inria,x}
  \end{icmlauthorlist}
  \icmlaffiliation{inria}{Inria Saclay}
  \icmlaffiliation{x}{LIX, CNRS, École Polytechnique, IP Paris}
  \icmlcorrespondingauthor{Julie Mordacq}{julie.mordacq@inria.fr}
  \icmlkeywords{Machine Learning, ICML}
  \vskip 0.3in
]



\printAffiliationsAndNotice{}  

\input{sections/00_abstract}
\input{sections/01_intro}
\input{sections/02_related-work}
\input{sections/03_method.tex}
\input{sections/04_xp.tex}
\input{sections/05_conclusion}

\section*{Acknowledgments}
This work is supported by Hi! PARIS, ANR/France 2030 program (ANR-23-IACL-0005)
and Inria Action Exploratoire {\sc PreMediT} (Precision Medicine using Topology).
We were granted access to the
HPC resources of IDRIS under the allocations 2025-A0190616899
made by GENCI.
We would like to thank David Loiseaux and Eleftherios Tsonis for their useful feedback.

\section*{Impact Statement}
This paper presents work whose goal 
is to advance the field of machine learning. 
There are many potential societal consequences of our work, 
none of which we feel must be specifically highlighted here.

\bibliography{citations}
\bibliographystyle{icml2026}

\newpage

\appendix
\onecolumn
\part*{\centering {\large Appendix to}\\
\name{}: Assessing Self-Supervised Learning Representations via Intrinsic
Dimension}

\let\addcontentsline\origaddcontentsline
\startcontents[appendix]
\printcontents[appendix]{}{1}{\section*{Contents}}

\input{sections/X0_dim_mst}
\input{sections/X2_pretraining}
\input{sections/XX_further_results.tex}

\input{sections/X3_compute-cost}

\input{sections/X4_further_benchmark.tex}
\input{sections/X5_effective_intrinsic.tex}


\end{document}

%% file: sections/00_abstract.tex
\begin{abstract}
Self-supervised learning (SSL) has emerged as a powerful paradigm for 
learning meaningful representations from unlabeled data. 
However, the standard protocol for evaluating these representations, 
linear probing,
is computationally expensive, 
sensitive to hyperparameters, and provides limited insight
into the geometric structure of the representation space.
In this work, motivated by connections between neural network generalization 
and intrinsic dimension (ID) we propose \name{}, 
a method for estimating the ID of SSL representations 
via the Minimum Spanning Tree dimension estimator ($\mathrm{dim}_\mathrm{MST}$).
Across diverse datasets, architectures, and SSL pretraining objectives, 
we show that \name{} strongly correlates with downstream linear probe
performances.
Furthermore, we demonstrate that \name{} enables efficient hyperparameter selection, 
significantly reducing the computational cost compared to supervised
alternatives.
Our results highlight intrinsic dimensionality 
as a principled geometric proxy for assessing SSL representations, 
complementing standard supervised probing protocols.
\end{abstract}

%% file: sections/01_intro.tex
\section{Introduction}

\input{figure/1_fms-intra}

\textit{Can the geometry of learned representations provide reliable, 
label-free insights into their quality for downstream tasks?}
Self-supervised learning (SSL) offers a natural setting to ask this question, as its objectives are explicitly 
designed to structure representations without access 
to labels~\cite{chen2020simple,bardes2022vicreg,assran2023self,
venkataramanan2025franca,simeoni2025dinov3,mordacq2025tregs}. 
Beyond their strong performance, 
SSL representations are valued for their ability to transfer 
to a wide range of downstream tasks with minimal 
supervision~\cite{dufour2024world80timestepsgenerative,couairon2025jafar,
maruani2025illustrator,degeorge2025far}.
Rather than optimize task-specific decision boundaries, 
SSL methods shape the organization of data in the representation space 
through geometric constraints such as alignment between views, 
feature uniformity, and variance control. 
%
As a result, intrinsic properties of the resulting manifold, e.g., 
curvature and spectral structure, have emerged 
as potential indicators of representation quality,
prompting recent efforts to investigate such geometric proxies~\cite{ansuini2019intrinsic,garrido2023rankme}.

Among them, intrinsic dimension (ID),
originally introduced by~\citet{bennett1969intrinsic},
has emerged as a particularly informative quantity:
it characterizes the effective number of degrees of freedom required 
to represent data in an embedding space.
%
Recent studies have revealed a monotonic relationship between a 
model's generalization error and the intrinsic dimension of 
its representations~\cite{ansuini2019intrinsic, konzeffect2024}, 
with lower intrinsic dimension often correlating with improved downstream 
accuracy.
These findings, grounded in the 
\textit{manifold hypothesis}~\cite{goodfellow2016deep},
suggest that representation quality is governed not only by separability,
but also by how efficiently information is compressed 
into low-dimensional geometric structure.
Yet, existing evidence is largely confined to supervised convolutional
networks, leaving open how intrinsic dimension behaves across modern
self-supervised methods and whether it indicates downstream
performance in this setting.

In practice, estimating ID is far from trivial. Intrinsic dimension admits
multiple mathematical formalizations, topological, fractal, and
information-theoretic, each capturing different facets of data geometry,
and since ID cannot be observed directly but must be estimated from finite
samples, the choice of estimator matters.
%
Most notably,
nearest-neighbor-based methods such as TwoNN~\cite{facco2017estimating} 
and maximum likelihood estimators~\cite{levina2004maximum} 
have been widely adopted, but suffer from well-known limitations: 
They rely on strong
locality and isotropy assumptions, are sensitive to noise and finite-sample
effects, and become unstable in high-dimensional or highly structured
representation spaces.
%
These limitations are especially pronounced in SSL, 
which requires operating far from standard conditions. 
First, it operates in a non-asymptotic regime 
where $n \approx d$ rather than $n \to \infty$ with $d$ fixed,
where $n$ is the number of samples and $d$ is the ambient dimension.
Second, SSL objectives introduce dependencies between data points, 
violating standard independence assumptions. 
For instance, as shown in \Cref{fig:assumptions-twonn}, 
TwoNN and MLE are sensitive to the latter; they are unable to reliably detect 
the intrinsic dimension, with TwoNN even diverging.


In this paper, we consider a complementary ID estimator grounded in the
scaling behavior of minimum spanning
trees~\cite{costaDeterminingIntrinsicDimension2006}.
Asymptotically tied to the intrinsic R\'enyi entropy, this estimator
balances local and global information, enjoying robustness to noise,
to sampling-density variations, and to high ambient dimensionality.
Furthermore, its reliability under sampling sparsity, a behavior
consistent with known theoretical results on MST length in sparse
regimes (e.g., Theorem~1 in~\cite{mordacq2025tregs}), makes it
particularly well-suited to the $n \approx d$ regime inherent in vision SSL.

Building on this perspective, we propose \name{} 
(pronounced as {\em id est}, Latin for `that is'), standing for \underline{ID} 
\underline{E}stimation for \underline{S}SL using minimum spanning \underline{T}ree.
\name{} is an unsupervised criterion for evaluating self-supervised representations 
based on intrinsic dimension estimation via minimum spanning trees.
We show that \name{} strongly correlates with downstream performance across a
wide range of self-supervised objectives, 
including joint-embedding, joint-predictive
and vision-language alignment
(\Cref{fig:fms-intra,fig:fms-inter,fig:fms-other-eval}). 
Moreover, \name{} provides an efficient 
proxy to supervised linear probing, enabling practical hyperparameter 
selection without labels at a fraction of the computational cost of supervised probing.

Our main contributions can be summarized as follows:
\begin{enumerate}[label=\textit{\roman*)},nosep]
    \item We propose \name{}, an unsupervised criterion based on the intrinsic dimension
    of the representation (\Cref{sec:method}).
    \item We demonstrate that \name{} serves as an efficient proxy to assess 
    the quality of SSL representations (\Cref{subsection:predict}).
    \item We show that \name{} enables unsupervised hyperparameter 
    selection across diverse SSL objectives and architectures 
    (\Cref{subsection:training-dynamics,subsection:model_selection}). 
\end{enumerate}

Beyond providing empirical insights, our work underscores the promise of geometric
descriptors as a complement to traditional supervised evaluation methods.

%% file: figure/1_fms-intra.tex
\begin{figure*}[t]
  \centering
    \includegraphics[width=\textwidth]{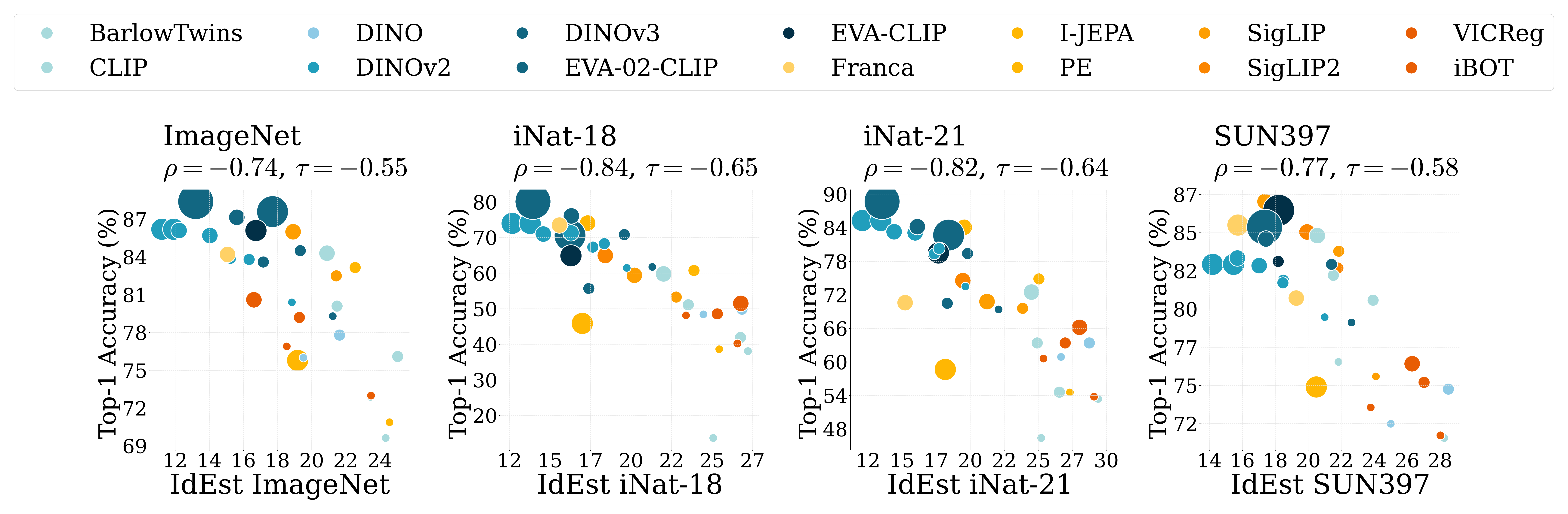}
  \caption{\textbf{Foundation Models and \name{}}: \hlamber{Intra-Dataset Correlation}. 
  Linear probe accuracy of pretrained SSL models 
  on 
  ImageNet (left), 
  iNat-18 (middle left),
  iNat-21 (middle right), SUN397 (right)
  versus \name{} 
  on each respective dataset.
  Each point corresponds to a model checkpoint; point size reflects the number of parameters.
  We report 
  Kendall's $\tau\in[-1,1]$ and Spearman's $\rho\in[-1,1]$.
  Correlations across all four benchmarks demonstrate 
  \name{}'s ability to provide insights into models' representation quality. 
  }
  \label{fig:fms-intra}
\end{figure*}

%% file: sections/02_related-work.tex
\section{Related Work}

Our work builds on recent efforts to analyze geometric properties 
of deep neural networks (DNNs) in a label-free setting, and to understand 
their connection to generalization. 
Existing literature investigates 
these properties through two main lenses: 
spectral properties and intrinsic dimensionality.


\subsection{Representation Spectrum}
$\alpha$-ReQ~\cite{alpha2022Agrawal} and RankMe~\cite{garrido2023rankme} 
characterize the eigenspectrum 
of representations within SSL frameworks, 
by measuring the decay rate of empirical eigenvalues or calculating the effective rank
of the representations, respectively. 
These studies showed that both metrics often strongly correlate 
with downstream performance and highlighted
their utility for hyperparameter selection.

However, the power-law assumptions underlying $\alpha$-ReQ 
are violated in the presence of representation 
collapse~\cite{he2022exploring} (where networks output identical or non-informative vectors regardless of the input, 
resulting in rank-deficient representations), 
leading to unreliable performance estimates when the 
embedding space becomes rank-deficient~\cite{garrido2023rankme}. 
Furthermore, RankMe is limited to the study of
Joint-Embedding Architectures (JEAs) where
two networks are trained to produce
similar embeddings for different views of the same image.
Notably, RankMe is tailored to identify a pivotal 
challenge of JEAs: representation collapse~\cite{jing2021understanding}.
This narrow focus potentially limits the applicability 
of such metrics to other SSL paradigms 
less prone to this specific type of collapse. 
For instance, as shown in \Cref{tab:hparam-selection}, 
RankMe is less effective on I-JEPA, a joint-predictive method.

More recently, \citet{thilaklidar} proposed LiDAR, 
which quantifies the rank of the Linear Discriminant Analysis 
matrix associated with the surrogate SSL task, a measure that intuitively 
captures how much information the representation retains for solving that task.
Crucially, LiDAR leverages SSL pretraining information (i.e., augmentations),
whereas our work targets the harder setting, where only frozen representations
are accessible, without any access to training data or pairing structure,
aiming to gain a deeper understanding of SSL models beyond hyperparameter selection.

\subsection{Intrinsic dimension}
Intrinsic dimension (ID) has been linked to the generalization 
of DNNs through two primary lenses: \textit{the optimization trajectory}, analyzing 
the sequence of model states and parameter updates during 
training~\cite{simsekli2020hausdorff,birdal2021intrinsic,dupuis2023generalization,
tan2024limitations}, and \textit{the learned representations}~\cite{ansuini2019intrinsic, 
konzeffect2024,ruppik2025less}.
In this work, we focus on the latter. Analyzing optimization trajectories 
is often computationally expensive, as it requires processing all network parameters 
(which can reach hundred  of millions in current vision models, e.g., 
ViT-L~\cite{dosovitskiy2021an}) 
and necessitates access to intermediate training checkpoints that are rarely available 
for large-scale, pre-trained models.

Regarding representational analysis, several studies have
investigated the ID of Large Language Model (LLM) representations~\cite{aghajanyan2021intrinsic,cai2021isotropy,
tulchinskii2023intrinsic,
valeriani2023geometry,
viswanathan2025geometry,lee2025shared,ruppik2025less}
demonstrating that ID provides critical insights into training dynamics and generalization.
However, research on computer vision models has largely remained 
restricted to supervised convolutional neural networks (CNNs). 
For instance, \citet{ansuini2019intrinsic} showed that the ID of 
supervised CNNs correlates with performance,
while \citet{konzeffect2024} proposed a generalization scaling law based on representational ID, 
though their validation was limited to supervised CNN architectures.

Despite these advances, the use of intrinsic dimension estimations to
characterize representation quality remains largely 
unexplored in self-supervised learning.



%% file: sections/03_method.tex
\section{Generalization and Dimension Estimation}\label{sec:method}

This section first introduces the theoretical connection between 
intrinsic dimension and generalization (\Cref{subsection:generalization}). 
Second, it presents standard intrinsic dimension estimators 
used in prior studies of deep neural network representations,  
along with their limitations, 
and motivates the use of an alternative: \dimmst{} (\Cref{subsection:id-estimators}).

\subsection{Theoretical connection to generalization}
\label{subsection:generalization}

Consider a classification dataset $\mathcal{D}$, consisting of $N_D$ points $x \in \mathbb{R}^{n}$ 
with target labels $y = \mathcal{F}(x)$ defined by an unknown function $\mathcal{F} : \mathbb{R}^{n} \to \mathbb{R}^C$, 
where $C$ is the number of classes. 
The dataset is split into a training set $\mathcal{D}_\text{train}$ and a test set $\mathcal{D}_\text{test}$. 
%
We analyze `well-trained' models $f : \mathbb{R}^{n} \to \mathbb{R}^C$ that interpolate the training data, 
such that $f(x) = \mathcal{F}(x)$ for all $x \in \mathcal{D}_\text{train}$. 
Let $\mathcal{L}$ be a non-negative loss function (e.g., cross-entropy) satisfying 
$\mathcal{L}(f(x), \mathcal{F}(x)) = 0$ if $f(x) = \mathcal{F}(x)$. 
The generalization error is expressed as the expected loss over $\mathcal{D}_\text{test}$. The model can
be decomposed as $f = h \circ g$, where $g$ is an encoder (e.g., a pre-trained SSL backbone) 
that produces latent \text{representations} living on some $d$-dimensional manifold, and where $h$ is a classification head.
%
\begin{theorem}{~\citet{konzeffect2024}}\label{thm:loss_bound}
 Let $K_L$ the Lipschitz constant of the loss function. Then:
    \begin{equation}
        \mathcal{L} \simeq \mathcal{O}\left(K_L N_D^{-1/d}\right)
     \end{equation}
\end{theorem}
This relation suggests that, for a fixed dataset size $N_D$, the error is dominated by the $-\frac{1}{d}$ exponent. 
Consequently, a lower $d$ implies a more efficient representation.
We therefore use the estimated ID of the representations as an unsupervised
criterion to assess the quality of the downstream task.


\subsection{Intrinsic dimension estimators}
\label{subsection:id-estimators}

The intrinsic dimension (ID) estimation problem assumes the data points are sampled on 
(or close to) some unknown $d$-submanifold of the ambient space.
The goal is to estimate $d$ from the data.

\begin{figure}
    \includegraphics[width=0.99\linewidth]{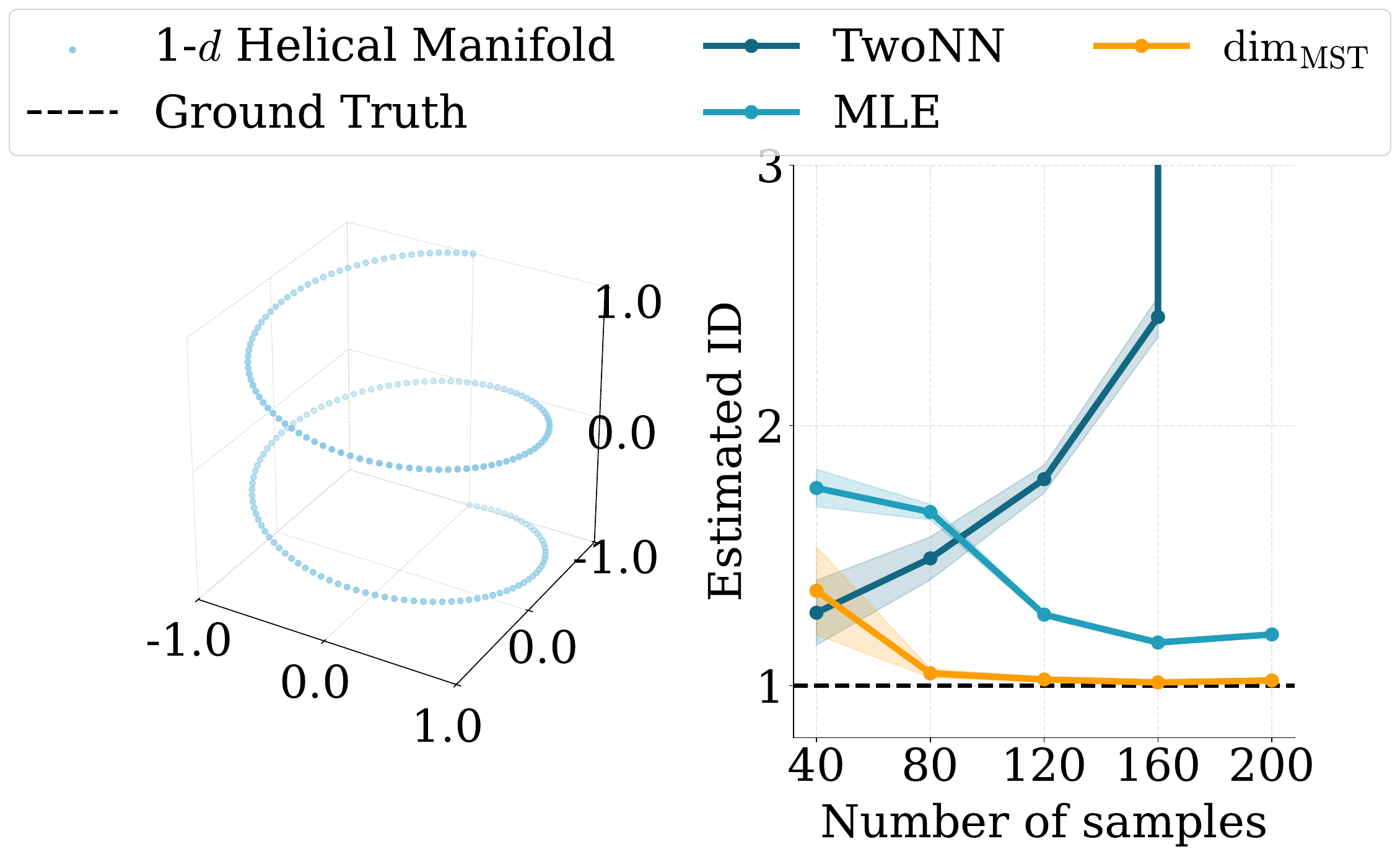}
    \captionsetup{font=normalsize}
    \caption{\textbf{Impact of Sampling Distribution on Estimators.} 
    \textit{(Left)} 
    200 points sampled evenly along a 1-dimensional helix. 
    \textit{(Right)} Estimated ID as a function of sample size. 
    While \dimmst{} 
    converges accurately to the ground truth $d=1$ as the sample size increases, 
    MLE and TwoNN do not, with TwoNN even diverging to infinity.
    }
    \label{fig:assumptions-twonn}
\end{figure}

\subsubsection{Parametric Estimators}
While many estimators exist~\cite{johnsson2014low,tempczyk2022lidl,binnie2025survey}, 
two main estimators 
have been adopted in prior studies of deep neural network 
representations~\cite{ansuini2019intrinsic, popeintrinsic, konzeffect2024, ruppik2025less}: 
\textit{\underline{Maximum Likelihood Estimation (MLE)}}~\cite{levina2004maximum} and
\textit{\underline{TwoNN}}~\cite{facco2017estimating}.
Both are parametric estimators, founded on the assumption that 
the data points are sampled i.i.d.
from a probability distribution supported on the submanifold, with locally constant density, 
and both treat the data points locally as a homogeneous Poisson process.
Under these conditions, the number of points within a small $\varepsilon$-ball, 
and the ratio between the distances to the second and first nearest neighbors, 
follow specific parametric distributions whose parameters depend directly on~$d$ and can be inferred, 
either via maximum likelihood estimation for MLE or via regression for TwoNN. 

These methods are inherently tied to specific data distributions, and they become unstable as the input data deviate from those, even in simple cases. 
Consider, for instance, the sample of \Cref{fig:assumptions-twonn}, 
composed of up to 200 points evenly spaced along a 1-dimensional helix embedded in $\mathbb{R}^3$. 
Despite the high regularity of both the submanifold and the sample,  both methods fail to recover
the correct intrinsic dimension, 
with TwoNN even diverging to infinity in the asymptotic regime. This behavior is consistent across standard choices of hyperparameters of the methods.
%
%
In practice, this instability translates into a degradation of the performances in the presence of noise \cite{tulchinskii2023intrinsic,binnie2025survey}, or on heavy-tailed distributions \cite{birdal2021intrinsic}.

\subsubsection{Metric Invariants Based Estimators}
The limitations of the parametric methods TwoNN and MLE motivate 
a shift toward estimators based on the theory of Euclidean 
functionals~\cite{yukich2006probability}.
In particular, the asymptotic growth rate of the
length of Minimum Spanning Tree (\mst{}) 
are related to the Rényi entropy of the underlying distribution.  
This connection has enabled the derivation of several 
dimension estimators with proven consistency 
under relatively weak assumptions:
compactness of the manifold and boundedness of the Lebesgue sampling density supported on the
manifold~\cite{costaDeterminingIntrinsicDimension2006}.
For instance, the {\em Minimum Spanning Tree dimension estimator}, 
or $\dimmst{}$ for short, 
successfully recovers the ground truth dimension $d=1$
in the example of~\Cref{fig:assumptions-twonn}.

Given a point cloud $Z$ in $\mathbb{R}^D$, the Minimum Spanning Tree (\mst{}) is
the acyclic connected graph $G=(V,E)$, with vertex set  
$V=Z$, that minimizes 
the total edge length:
\begin{equation}
    L(G) = \sum_{(z,z')\in E} \|z-z'\|_2.
\end{equation}
%
\citet{costaDeterminingIntrinsicDimension2006} studied the growth rate of the length of the minimum spanning tree for random point clouds in Riemannian manifolds.
%
%
Given an i.i.d. 
$n$-sample $X_n$ drawn from a fixed probability measure $P_X$  supported on a compact Riemannian $d$-manifold~$M$ with density $f_X$ w.r.t. the Hausdorff measure, there exists a constant $C'$ independent of $f_X$ and of $M$ such that, almost surely:
	\begin{equation}
		n^{-\frac {d-1}{d}} \cdot \costmst{X_n} \xrightarrow[n\to \infty]{}{C' \int f_X^{\frac{d-1}{d}}} \mathrm{d}\mathcal{H},
		\label{eq:steelecosta} 
    \end{equation}
where $\mathcal{H}$ denotes the Hausdorff measure on~$M$.

This result motivates the definition of \dimmst{} given 
by~\citet{costaDeterminingIntrinsicDimension2006}:
\begin{definition}
    Given a bounded metric space~$M$, the \mst{} dimension of $M$, 
    denoted by $\mathrm{dim}_{\mst{}}(M)$, 
    is the infimal exponent
    $d\in \mathbb{N}$ such that $\costmst{X}/|X|^{\frac {d-1} d}$ 
    is uniformly bounded for all finite subsets $X \subseteq M$:
\begin{equation*}
\begin{split}
 \dimmst{}(M) := \mathrm{inf}\Big\{d : \exists C \text{ s.t. } \frac{\costst{\mathrm{MST}(X)}}{|X|^{\frac {d-1} d}} \leq C \, \\
 \textnormal{ for every finite subset }X \textnormal{ of }M\Big\}.
 \end{split}
\label{eq:dim_mst}
\end{equation*}
\end{definition}

In practice, the ID is estimated via log-log linear 
regression~\cite{birdal2021intrinsic, binnie2025survey}. 
Specifically, given subsamples $X_{n_i}$ with increasing sizes $n_i$, we fit:
\begin{equation*}
        \log(L(\mst{}(X_{n_i}))) \approx \frac{d-1}{d} \log(n_i) + \log(C).
\end{equation*}
The resulting slope $m$ yields the intrinsic dimension via the relation $d = 1/(1-m)$.
The complete algorithm for computing $\dimmst{}$ is given in~\cref{alg:dimmst}.

\paragraph{Persistent Homology Dimension.}
In Topological Data Analysis (TDA), 
the \mst{} relates to the so-called \emph{total degree-0 persistence of the Rips filtration}~\cite{oudot2015persistence}. 
This connection allows for the definition of the \emph{0-dimensional Persistent Homology (PH) dimension}, 
$\text{dim}_\text{PH}^0$, which is equivalent 
to the \dimmst{}~\cite{adamsFractalDimensionMeasures2020}.
The PH dimension has been used in several dimension-estimation 
applications~\cite{birdal2021intrinsic,dupuis2023generalization}.
This connection provides further theoretical grounding for the robustness of \dimmst{}. 
In particular, TDA-based
measures such as the total persistence of the Rips 
filtration are provably stable under pertubations of 
the underlying distribution~\cite{chazal2014persistence}.
This was further observed empirically by~\citet{tulchinskii2023intrinsic}
 in the context of LLM latent spaces.

%% file: sections/04_xp.tex
\section{\name{}: \dimmst{} for Unsupervised Assessment of Self-Supervised methods}

In this section, we apply $\dimmst{}$ to estimate 
the intrinsic dimension of self-supervised representations. 
This yields \name{}, which 
stands for \underline{I}ntrinsic \underline{D}imension \underline{E}stimation 
for SSL using minimum spanning \underline{T}rees. 
Our goal is to determine whether \name{} can yield unsupervised
insights into downstream performances.
Specifically, to empirically validate \name{}, 
we compare it against linear probing, 
the standard evaluation protocol for self-supervised learning (SSL). 


Subsequently, we address the following research questions: 
\begin{enumerate}[label=\textit{Q\arabic*.}, leftmargin=*, nosep]
  \item To what extent does \name{} reflect 
  representation quality across pretrained SSL models? 
  \item Can \name{} yield insight along self-supervised pretraining?
  \item Can \name{} be leveraged as a principled proxy for hyperparameter 
  selection without requiring labeled data?
\end{enumerate}

\subsection*{Implementation of \name{}}

To satisfy the assumptions of~\Cref{thm:loss_bound}, \name{} operates on 
the frozen representation 
passed to the classifier head, following each method's standard evaluation 
protocol (i.e., consistent with how linear probes are trained). 
For models without a class token (e.g., I-JEPA), 
we average-pool the patch 
tokens; for models with a \texttt{[CLS]} token (e.g., DINO, DINOv2), 
we use this token directly.
Additional implementation details are provided 
in Appendix~\ref{sec:supmat-idest-implementation}.

\subsection{\name{} and accuracy of foundation models}
\label{subsection:predict}

\input{figure/1_fms-further.tex}
\input{figure/1_fms_eval}

We begin by evaluating whether our proposed \name{} 
reflects representation quality across a diverse set of pretrained SSL models. 
Specifically, we compute \name{} on frozen representations 
and compare it against standard linear probing accuracy. 
We compute two rank correlation statistics:
\textit{Spearman's rank correlation coefficient} $\rho$~\cite{spearman1961proof}
and \textit{Kendall's rank correlation coefficient} $\tau$~\cite{kendall1938new}.
Both capture monotonic relationships between rankings.


\noindent \textbf{Setup.} 
We evaluate 14 SSL methods spanning four paradigms: 
pure joint-embedding (e.g., VICReg~\cite{bardes2022vicreg}, 
DINO~\cite{caron2021emerging}), 
joint-predictive (e.g., I-JEPA~\cite{assran2023self}), 
combined objectives (e.g., iBOT~\cite{zhou2021ibot}, 
DINOv2~\cite{oquab2023dinov2}), 
and vision-language alignment (e.g., CLIP~\cite{radford2021learning}, 
EVA-CLIP~\cite{sun2023eva}).

%
For each method, we evaluate two main architectures: 
ResNet~\cite{he2016deep} and ViT~\cite{dosovitskiy2021an}, 
and we include various model scales (e.g., ViT-S, and ViT-G). 
This results in 33 different models. 
\Cref{tab:model-used} (in supplementary) 
provides the complete list of models and architectures studied.

\noindent\textbf{Results.}
We first examine \hlamber{Intra-Dataset Correlation} in \Cref{fig:fms-intra}, 
where \name{} is estimated on each dataset separately 
and compared against the corresponding 
linear probe accuracy.
\Cref{fig:fms-intra} reports results on ImageNet~\cite{deng2009imagenet}
and three additional fine-grained datasets: iNat-18~\cite{van2018inaturalist}, 
iNat-21~\cite{van2021benchmarking}, 
and SUN397~\cite{xiao2010sun}.
Each point represents a pretrained SSL model, spanning diverse architectures,
training objectives, and scales.

Across all datasets, \Cref{fig:fms-intra} reveals a consistent
negative correlation between \name{}
and downstream linear probing accuracy:
models with lower intrinsic dimension tend to achieve higher performance.
This trend holds on ImageNet (left) as well as on the fine-grained benchmarks
iNat-18 (middle left), iNat-21 (middle right), and SUN397 (right). 
Notably, the relationship is consistent across a wide range of SSL paradigms,
joint-embedding methods, joint-predictive methods, 
and vision-language pretraining,
suggesting that intrinsic dimension acts as a unifying geometric descriptor
of representation quality, independent of the training objective.

This is further supported by the correlation metrics reported atop each plot.
Both Kendall's $\tau \approx -0.6$ and Spearman's $\rho \approx -0.8$ confirm that
\name{} reliably preserves the relative ordering of models across all four datasets,
in strong agreement with linear probing rankings.


While our analysis has first focused on Intra-Dataset Correlation,
a natural question arises: do these findings persist when \name{} and accuracy
are measured on \emph{different} datasets, 
or under alternative evaluation protocols?

In~\Cref{fig:fms-inter}, we study \hlskyblue{Inter-Dataset Correlation},  
where \name{} is computed on ImageNet and accuracy is evaluated on
four target datasets: the large-scale fine-grained benchmarks iNat-18 and iNat-21,
and the smaller-scale datasets CIFAR-10 and CIFAR-100. 
The correlation remains strong across both target datasets,
suggesting that \name{} captures intrinsic 
properties of the learned representations
rather than dataset-specific characteristics.

In~\Cref{fig:fms-other-eval}, we examine \name{} 
under \hlcerulean{Alternative Evaluation Protocols}.
\name{} remains indicative of performance 
on ImageNet under kNN evaluation,
and on the complementary ImageNet-v2 validation set.

Overall, these results reveal that 
low intrinsic dimension is a 
consistent feature of well-performing representations, 
and suggest that \name{} serves as a simple, 
label-free proxy for downstream evaluation 
(\Cref{fig:fms-intra,fig:fms-inter,fig:fms-other-eval}), 
offering insight into the quality of self-supervised representations 
without annotated data.
\vspace{3mm}
\hypbox[Finding 1.]{
    \textbf{\name{} negatively correlates 
    with linear probing accuracy:} 
    low intrinsic dimension is a consistent geometric 
    signature of strong representations
    across intra- and inter-dataset settings
    and alternative evaluation protocols.
}


%

\subsection{Training dynamics: offline and online probing}
\label{subsection:training-dynamics}

\input{figure/2_training-dynamics.tex}




Here, we study whether \name{} tracks representation quality during unsupervised training. 
To this end, we consider two complementary evaluation protocols: 
\textit{(i) offline linear probing} and \textit{(ii) online probing}. 
While both aim to assess downstream performance over the course of self-supervised pretraining, 
they differ in when and how the classifier is trained.

\noindent\textbf{Offline linear probing.}
Offline linear probing is the standard evaluation protocol in self-supervised learning. 
The representation model is first trained without labels, 
after which a linear classifier is trained on frozen features. 
To analyze training dynamics, we extract multiple checkpoints 
during pretraining and perform linear probing independently for each checkpoint.

\Cref{fig:training-dynamics} (top) reports results for VICReg~\cite{bardes2022vicreg} 
(\ref{fig:vicreg-offline-training}), DINO~\cite{caron2021emerging} (\ref{fig:dino-offline-training}) 
and I-JEPA~\cite{assran2023self} (\ref{fig:ijepa-offline-training}), 
respectively on ImageNet. 
We show linear probing top-1 accuracy (y axis on the left, 
in \textcolor{tigerorange}{orange}) as a function of training epochs, 
together with \name{} (y axis on the right, 
in \textcolor{cerulean}{blue}). 
Across both models, \name{} closely follows the evolution 
of downstream accuracy throughout training. 
As representations improve and linear probing accuracy increases, 
the intrinsic dimension consistently decreases. 
This strong temporal correlation indicates that \name{} 
captures meaningful geometric changes in the representation space as 
training progresses, without requiring labels.


\noindent\textbf{Online linear probing.}
We consider online probing, where a linear classification 
head is attached to the representation and trained jointly 
during self-supervised pretraining. 
Importantly, gradients from the classifier 
do not backpropagate into the representation encoder, 
ensuring that the learned features remain purely self-supervised. 
This setting allows us to monitor downstream 
performance continuously during training.

\Cref{fig:training-dynamics} (bottom) 
illustrates the training dynamics 
for DINO using a ViT-S, reporting the self-supervised loss (left), 
online classification loss and ImageNet top-1 accuracy (middle), 
and \name{} (right). We observe that during the early stages of 
training, particularly within the first 10 epochs, representations 
are highly constrained, rendering \name{} less informative. 
However, as training progresses, the self-supervised 
loss decreases and classification accuracy improves; concurrently, 
\name{} systematically drops, closely tracking the evolution 
of the other metrics. This demonstrates that \name{} faithfully 
reflects representation quality even in an online setting, capturing 
improvements in downstream performance as they emerge during the pretraining.


Overall, we observe that across both offline and online protocols, 
and across diverse SSL objectives, \name{} mirrors the 
evolution of downstream classification performance. 
These results further support its utility as a label-free indicator 
of representation quality, effective not only at convergence 
but throughout the pretraining process, once the initial training stages are surpassed. 
This leads us to our second finding:

\hypbox[Finding 2.]{
    %
    \textbf{\name{} can serve as a label-free proxy for SSL pretraining effectiveness:} 
    it decreases as downstream task accuracy improves over 
    the course of unsupervised pre-training.
}

\subsection{Label-free metric for Model Selection}
\label{subsection:model_selection}
\input{tables/hparam-selection}

As stated in \Cref{thm:loss_bound} and highlighted by previous experiments, 
performance on downstream tasks is bounded by the intrinsic dimension 
of the representations. 
Consequently, we evaluate \name{} as an hyperparameter 
selection criterion that may bypass the need for linear probing.

\noindent\textbf{Selecting hyperparameters with \name{}.}
\label{subsection:hparam}
Given a set of candidate models $\mathcal{F} = \{f_1, \dots, f_n\}$, 
each trained with varying hyperparameters,
and their corresponding dimension estimates $\Delta = \{d_1, \dots, d_n\}$, 
the selected model, $f^*$, is defined as: 
\begin{equation}
    f^* = f_{\arg\min_{i} d_i}
    \label{eq:select-hparam}
\end{equation}

\noindent \textbf{Set-Up.} 
We apply~\Cref{eq:select-hparam}
to identify the optimal model for a given hyperparameter configuration 
across several SSL frameworks including Joint-Embedding methods 
(VICReg~\cite{bardes2022vicreg}, DINO~\cite{caron2021emerging})
and the Joint-Predictive architecture (I-JEPA~\cite{assran2023self}).
Training details for each model are provided in Appendix~\ref{subsec:hparam-models}.

We focus on various hyperparameters: 
\begin{enumerate}[nosep]
    \item Optimization: learning rate (\texttt{lr}) and weight decay (\texttt{wd});
    \item Loss-specific coefficients: variance coefficients in VICReg (\texttt{var.});  
    the teacher and student temperatures in DINO (\texttt{t-temp.}, \texttt{s-temp.});
    \item Size of masking for I-JEPA: target block size (\texttt{target-size}) 
    and context block size (\texttt{context-size}).
\end{enumerate}

We evaluate performance on ImageNet and the average accuracy
across several fine-grained classification datasets,
comparing models selected by \name{} against those selected
by supervised validation accuracy, and include the \texttt{\textbf{all}} setting,
where selection is performed over the full hyperparameter pool, the
more challenging configuration.

\noindent\textbf{Results.}
As shown in~\Cref{tab:hparam-selection}, 
\name{} can recover most ImageNet oracle performance, 
achieving results near the upper bound of 
validation accuracy across various architectures (ResNets and ViTs) and pre-training objectives.
Notably, \name{} outperforms $\alpha$-ReQ in most settings without suffering 
from significant performance drops in the worst-case scenarios. 
For instance, $\alpha$-ReQ selects the lower-bound accuracy for I-JEPA.

Additionally, while RankMe was originally designed for JE-SSL methods (e.g., VICReg, DINO), 
and in fact mirrors the VICReg objective by directly rewarding high effective rank, 
\name{} maintains competitive performance on these models 
while also generalising to settings where RankMe struggles, such as I-JEPA.
%
\name{} demonstrates versatility to other SSL paradigms.
As I-JEPA is not a standard Joint-Embedding Method 
(making dimensional collapse less of a primary concern), 
other metrics struggle, whereas \name{} 
successfully generalizes to this and other SSL paradigms.

Furthermore, LiDAR~\cite{thilaklidar} is a strong baseline when pretraining augmentations are
accessible.
Nonetheless, across all four models in the \texttt{\textbf{all}} setting,
\name{} achieves comparable model selection than LiDAR
in a strictly unsupervised setting.
Additional results on DINO (ResNet-50) are available in~\Cref{tab:hparam-dino-resnet}.

\vspace{3mm}
\hypbox[Finding 3.]{
    \textbf{\name{} can serve as an efficient label-free criterion for hyperparameter selection:}
    it performs comparably to supervised baselines across diverse architectures and SSL objectives.
}

\subsection{Computational Cost}
\input{tables/compute}

We evaluate the computational cost of \name{} and compare it to linear probing.
Compared to training a linear probe, which requires multiple epochs over
the training dataset features to obtain reasonable estimates of downstream
accuracy, computing \name{} only requires feature extraction
and a \dimmst{} computation.
In~\Cref{tab:compute}, we compare the wall-clock compute time for \name{} 
and standard linear probe training
on ImageNet (10 epochs across 2 GPUs), averaged
over 3 runs. Computations were performed on H100 GPUs and an Intel Sapphire Rapids 8468 CPU.
Across all architectures, \name{} is substantially faster than a 10-epoch 
linear probe ($B=1024$, on 2 GPUs). 
Notably, for larger models, the cost is dominated 
by feature extraction, while the overhead of computing \dimmst{} 
remains minimal across various output dimensions.

\subsection{Limitations}

While versatile, our framework has several limitations.

The formula in~\Cref{thm:loss_bound}, 
provides only a bound on the convergence rate. 
Therefore, even if two models have similar intrinsic dimensions, 
their actual convergence rates may potentially differ in practice. 
Thus, \name{} should be viewed more as an indicator of accuracy 
than as a perfect predictor. This is indeed reflected 
in~\Cref{fig:fms-intra} for instance, 
where $\rho \approx 0.8, \tau \approx 0.6$
confirms strong global ranking ability, while
acknowledging that the intrinsic dimension 
might not explain all the variance.
In particular, 
ranking is inherently harder when model accuracies are tightly 
clustered, as in ImageNet or SUN397, than when they are more spread out, 
as in iNaturalist.
Furthermore, one family deviates most visibly:
vision-language models (e.g., CLIP, EVA-CLIP), 
where the cone effect introduces geometric misalignment 
between encoders that the contrastive objective preserves 
rather than resolves, constraining the representations~\cite{liang2022mind}.

Additionally, a more subtle limitation stems from the early-training regime of ViTs: \name{} is less 
informative during the first 10 training epochs, 
before representations develop 
stable geometric structure (\Cref{subsection:training-dynamics}).

%% file: figure/1_fms-further.tex
\begin{figure}[t]
    \centering
    \begin{subfigure}{\linewidth}
      \begin{subfigure}{\linewidth}
        \includegraphics[width=\linewidth]{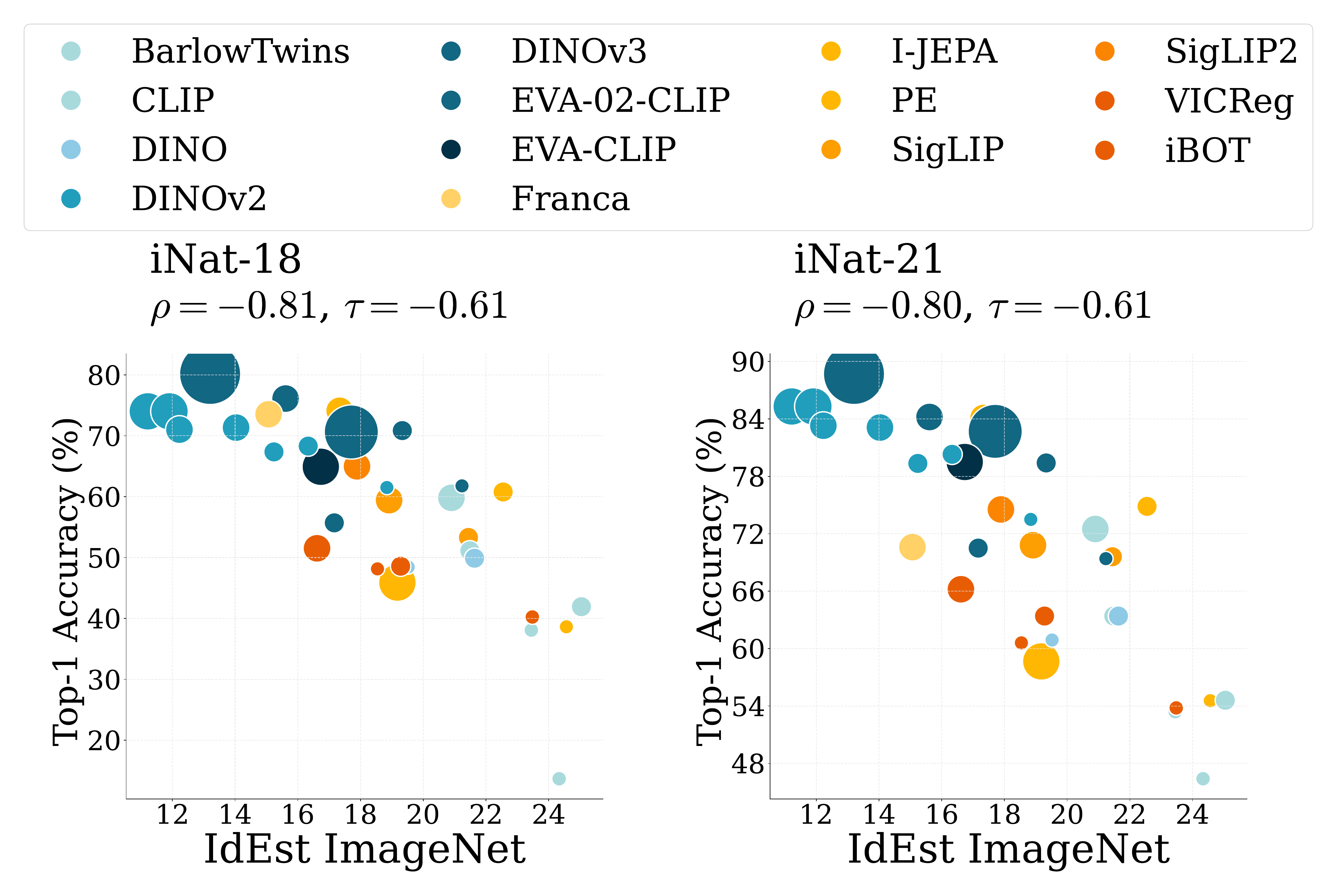}
      \end{subfigure}
      \vfill
      \vspace{0.2cm}
      \begin{subfigure}{\linewidth}
        \includegraphics[width=\linewidth, trim=0 0 0 12.5cm, clip]{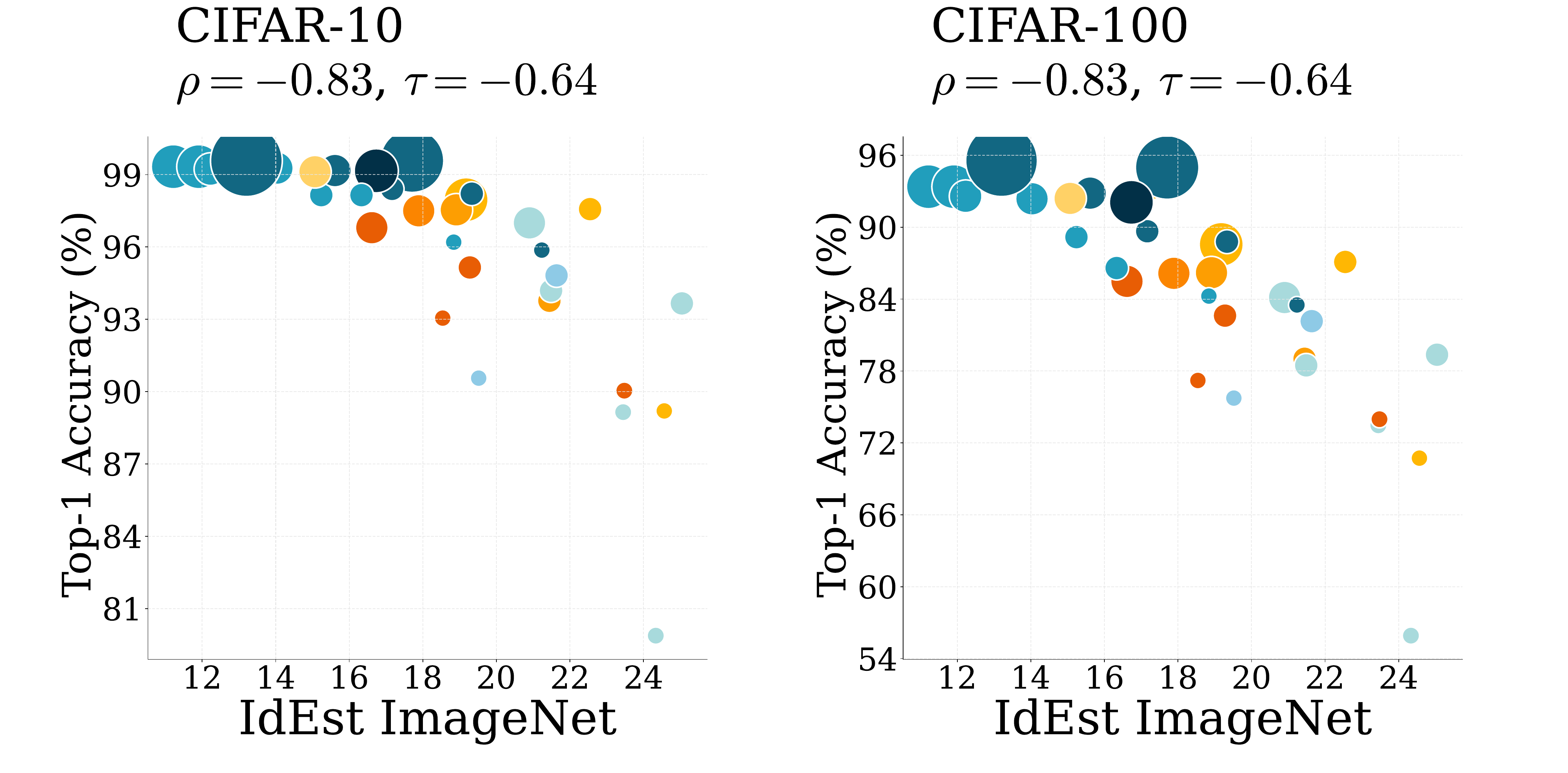}
      \end{subfigure}
    \end{subfigure}
    \caption{\textbf{Foundation Models and \name{}:} \hlskyblue{Inter-Dataset Correlation}.
    Linear probe accuracy of pretrained SSL models 
    on iNat-18 (top right), iNat-21 (top left), CIFAR-10 (bottom right), 
    CIFAR-100 (bottom left) versus \name{} computed on ImageNet.
    Strong correlations demonstrate that \name{} computed 
    on a single reference dataset: ImageNet, is indicative of
    model quality across datasets.
    }
    \label{fig:fms-inter}
  \end{figure}

%% file: figure/1_fms_eval.tex
\begin{figure}[t]
        \includegraphics[width=\linewidth]{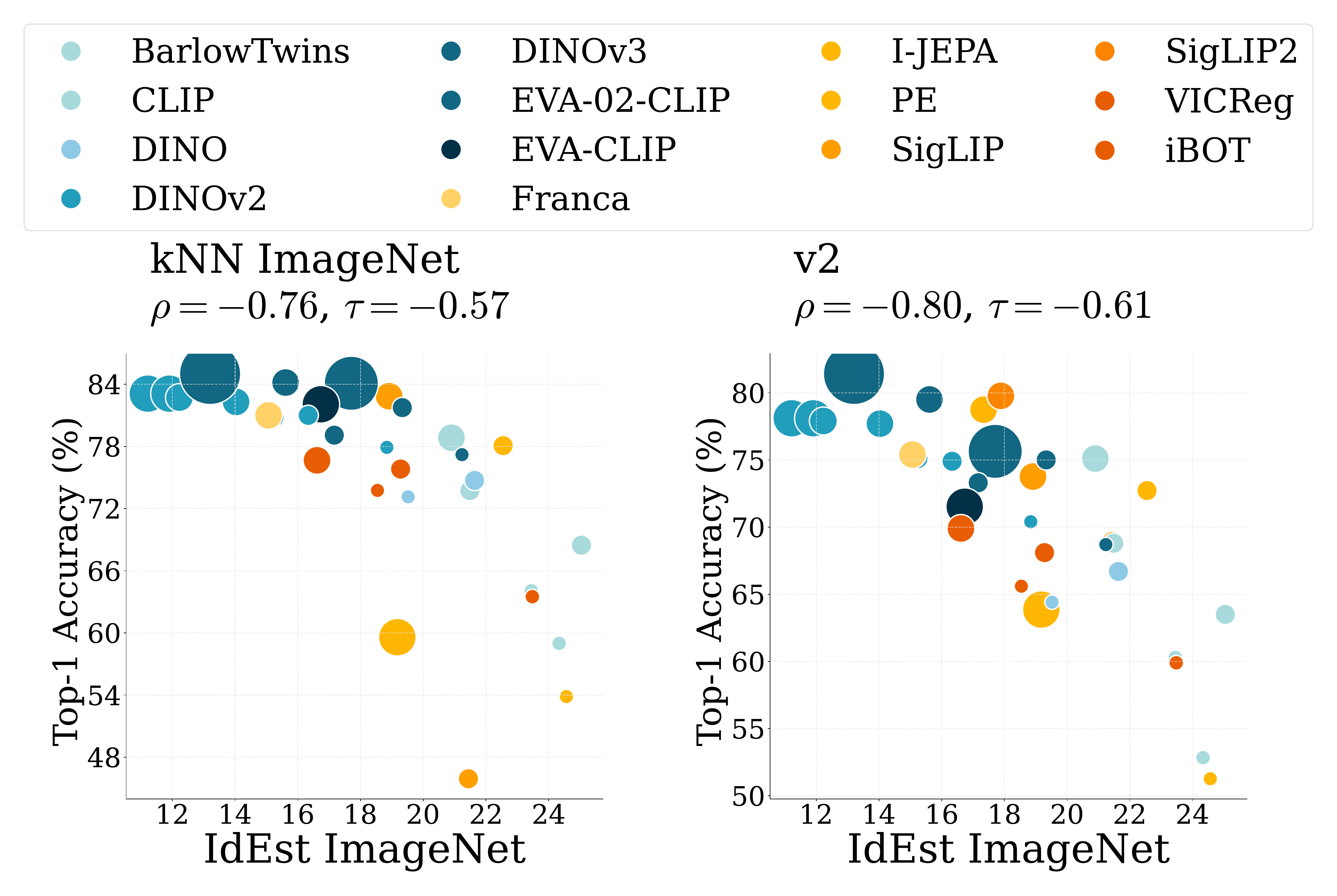}
    \caption{\textbf{Foundation Models and \name{}:} \hlcerulean{Alternative Evaluation Protocol}.
    Accuracy under alternative evaluation 
    settings versus \name{}.
    Strong correlations demonstrate that \name{} computed 
    on a single reference dataset: ImageNet, is indicative of
    model quality across evaluation protocols.
    }
    \label{fig:fms-other-eval}
  \end{figure}

%% file: figure/2_training-dynamics.tex
\begin{figure*}[t]
    \centering
    \begin{subfigure}{0.95\textwidth}
        \begin{subfigure}{0.32\linewidth}
            \centering
            \includegraphics[width=\textwidth]{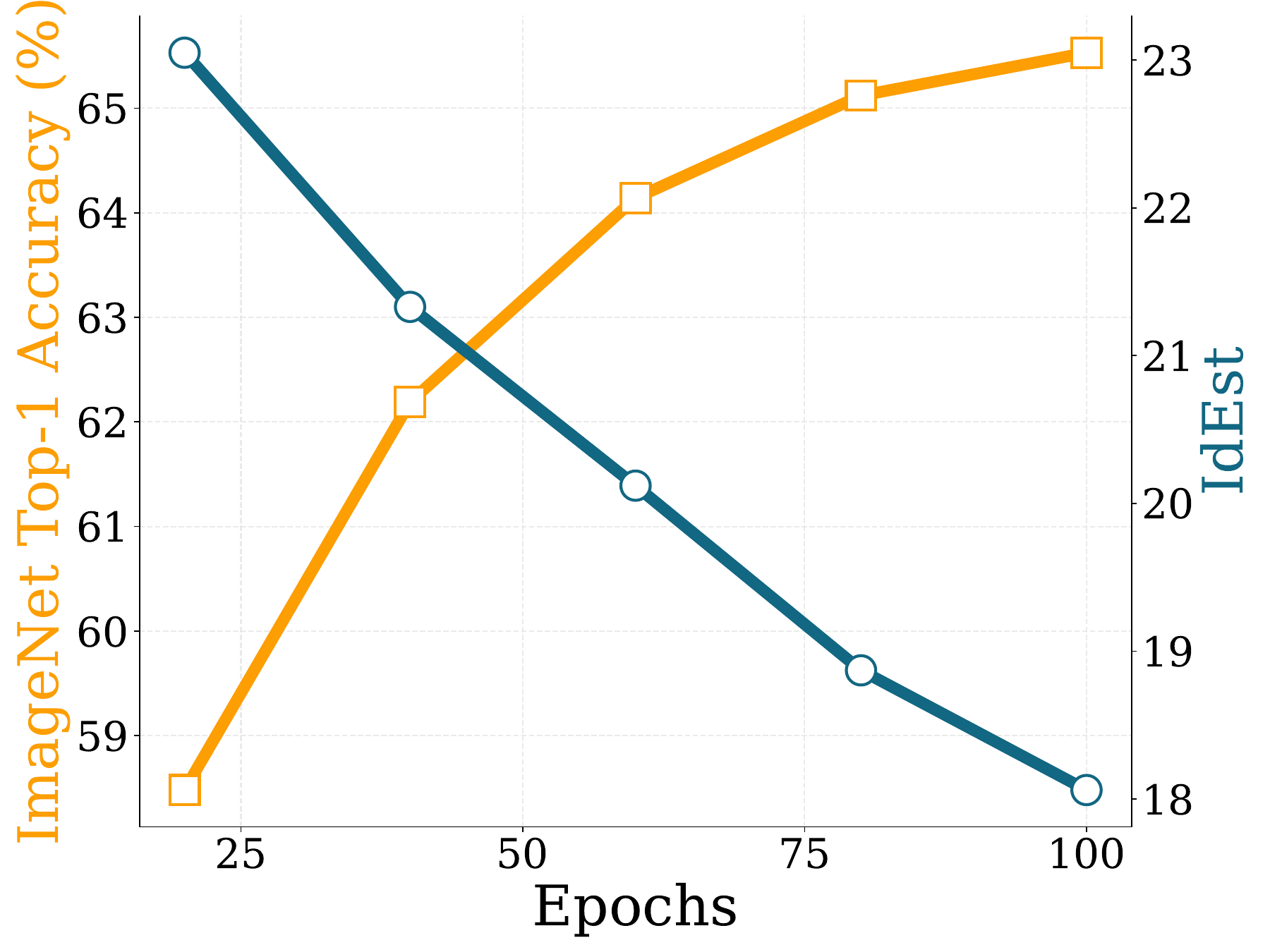}
            \caption{VICReg}
            \label{fig:vicreg-offline-training}
        \end{subfigure}
        \hfill
        \begin{subfigure}{0.32\linewidth}
            \centering
            \includegraphics[width=\textwidth]{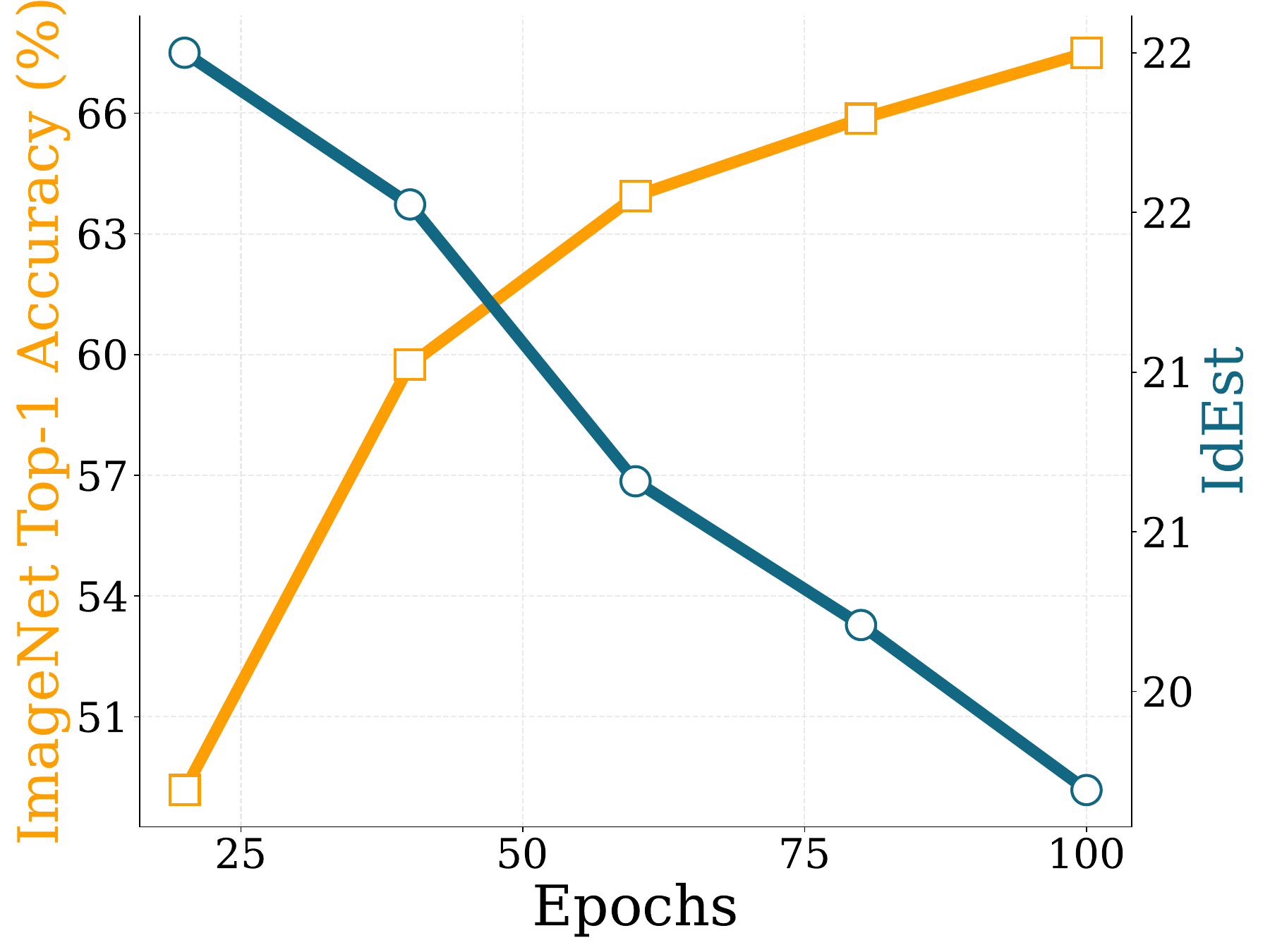}
            \caption{DINO}
            \label{fig:dino-offline-training}
        \end{subfigure}
        \hfill
        \begin{subfigure}{0.32\linewidth}
            \centering
            \includegraphics[width=\textwidth]{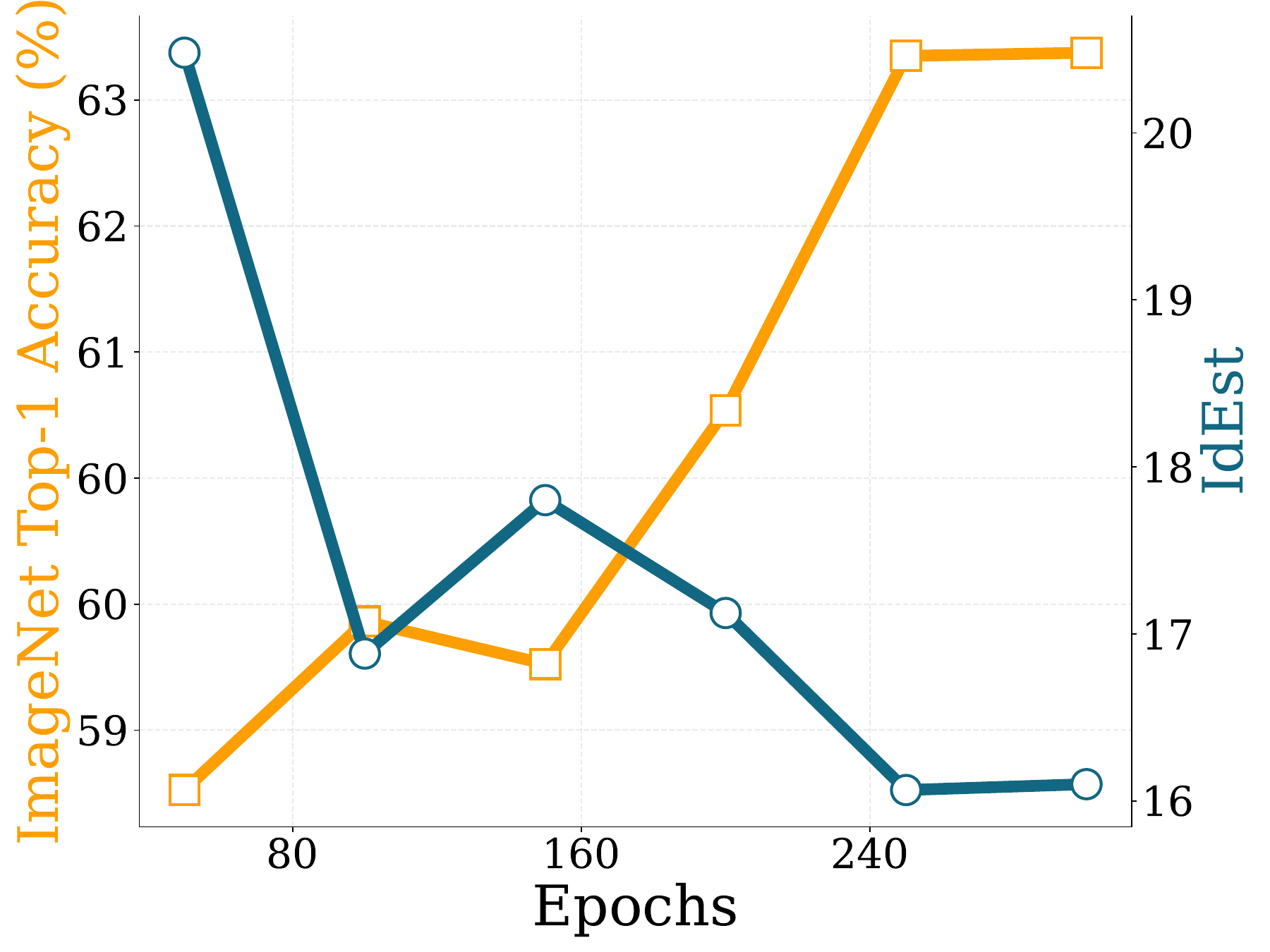}
            \caption{I-JEPA}
            \label{fig:ijepa-offline-training}
        \end{subfigure}
        \caption*{\underline{Offline probing dynamics.}}
        
    \end{subfigure}
    \begin{subfigure}{0.95\textwidth}
        \includegraphics[width=\textwidth]{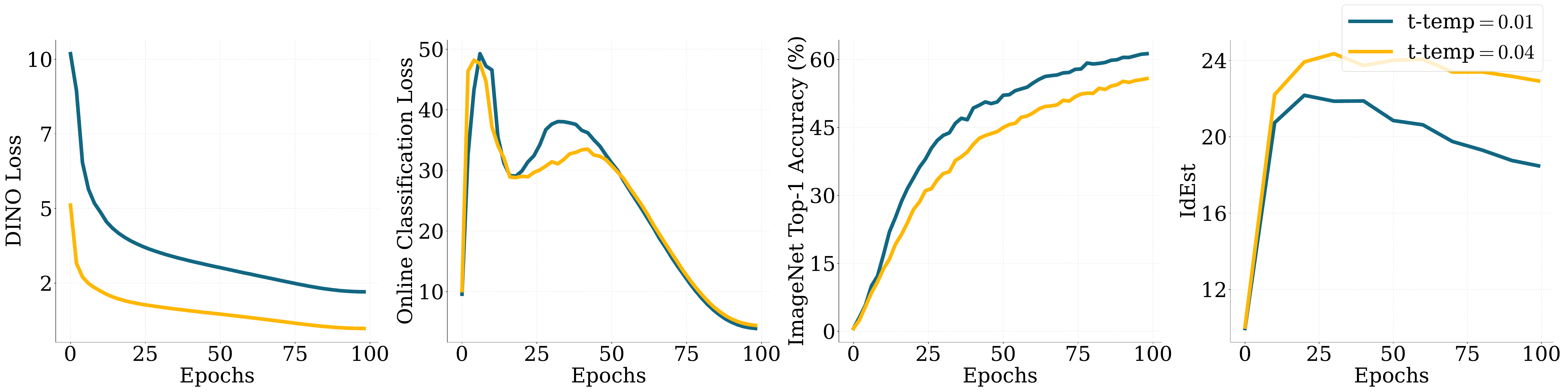}
        \refstepcounter{subfigure}
        \label{fig:dino-training}
        \caption*{\underline{Online probing dynamics during DINO pretraining.}}
       
    \end{subfigure}
    \caption{
    \textbf{Tracking Training Dynamics.}
    \textit{(Top)  Offline probing dynamics:} Evolution of ImageNet-1k 
    linear probing top-1 accuracy (left y-axis, 
    in \textcolor{tigerorange}{orange}) and \name{} 
    (right y-axis, in \textcolor{cerulean}{blue}) over 
    self-supervised pretraining epochs for (a) VICReg (ResNet-50), 
    (b) DINO (ViT-S), and (c) I-JEPA (ViT-B). 
    As representations improve and linear probing accuracy increases, 
    \name{} consistently decreases, demonstrating 
    its ability  to capture meaningful 
    geometric changes during the evolution of the representation.
    \textit{(Bottom) Online probing dynamics during self-supervised training:}
     Evolution of the self-supervised loss, classification loss, 
     ImageNet-1k online classification top-1 accuracy, and \name{} 
     over training epochs for DINO. While early-stage representations 
     are highly constrained, \name{} progressively tracks 
     improvements in downstream accuracy as training proceeds.
    }
    \label{fig:training-dynamics}
\end{figure*}

%% file: tables/hparam-selection.tex
\begin{table*}[t]
    \centering
    \renewcommand{\arraystretch}{1.2}
    \resizebox{0.99\linewidth}{!}{
    \begin{tabular}{
        ccc c cccc c cccc c cccc
    }
        \toprule
        &  &  & & \multicolumn{4}{c}{\textbf{VICReg} (RN-50)} & 
        & \multicolumn{4}{c}{\textbf{DINO} (ViT-S)} &
        & \multicolumn{4}{c}{\textbf{I-JEPA} (ViT-B)} \\
        \cmidrule(lr){5-8} \cmidrule(lr){10-13} \cmidrule(lr){15-18}
        
        \multirow{2}{*}{Dataset} & \multirow{2}{*}{Method} & \multirow{2}{*}{\rotatebox[origin=c]{90}{Unsup.}} & 
        & \multirow{2}{*}{\texttt{lr}} & \multirow{2}{*}{\texttt{wd}} & \multirow{2}{*}{\texttt{var.}}
        & \multirow{2}{*}{\textbf{\texttt{all}}} & 
        & \multirow{2}{*}{\texttt{lr}} & \multirow{2}{*}{\texttt{s-temp.}} &
        \multirow{2}{*}{\texttt{t-temp.}} & \multirow{2}{*}{\textbf{\texttt{all}}} &  
        & \multirow{2}{*}{\texttt{lr}} & \texttt{target-} & \texttt{context-} 
        & \multirow{2}{*}{\textbf{\texttt{all}}} \\
         &  &  & 
        & & &  &  &  &  &  &  &  & 
        &  & \texttt{size} & \texttt{size} &  \\

        
        \midrule
        \multirow{5}{*}{\rotatebox[origin=c]{90}{ImageNet}} 
        & \grayout{ACC-1 Bounds} &
        & & \grayout{[62.2, 66.5]} & \grayout{[37.5, 69.1]} & \grayout{[65.5, 67.2]} & \grayout{[37.5, 69.1]}
        & & \grayout{[63.6, 69.1]} & \grayout{[48.4,67.8]} & \grayout{[61.2 ,67.5]} & \grayout{[48.4, 69.1]}
        & & \grayout{[61.9, 66.4]} & \grayout{[49.0, 66.4]} & \grayout{[61.3, 66.5]} & \grayout{[49.0,66.5]} \\
        & $\alpha$-ReQ & \grayout{\checkmark}
        & & 65.0 & 53.5 & \best{66.9} & 53.5
        & & 63.6 & 58.6 & 61.2 & 58.6
        & & 61.9 & 49.0 & 61.3 & 49.0 \\
        & RankMe & \grayout{\checkmark}
        & & \best{66.5} & \best{69.1} & 65.5 & \best{69.1}
        & & 63.6 & \best{67.8} & \best{67.5} & 63.6
        & & 61.9 & 55.9 & 61.3 & 61.9 \\
        & LiDAR & \grayout{\texttimes}
        & & 65.0 & \best{69.1} & 66.8 & 65.0
        & & \best{68.2} & 65.6 & 65.0 & \best{65.5}
        & & 63.4 & \best{66.4} & \best{66.0} & 63.4 \\
        & \cellcolor{skylightblue!35}\name{} & \cellcolor{skylightblue!35}\grayout{\checkmark}
        & \cellcolor{skylightblue!35} & \cellcolor{skylightblue!35}62.3 & \cellcolor{skylightblue!35}67.1 & \cellcolor{skylightblue!35}65.5 & \cellcolor{skylightblue!35} 65.5
        & \cellcolor{skylightblue!35} & \cellcolor{skylightblue!35}64.7 & \cellcolor{skylightblue!35}65.6 & \cellcolor{skylightblue!35}\best{67.5} & \cellcolor{skylightblue!35}\best{65.5}
        & \cellcolor{skylightblue!35} & \cellcolor{skylightblue!35}\best{66.4} & \cellcolor{skylightblue!35}\best{66.4} & \cellcolor{skylightblue!35}\best{66.0} & \cellcolor{skylightblue!35}\best{66.4} \\
        \midrule
        \multirow{5}{*}{\parbox[c][10.5ex][c]{2mm}{\rotatebox[origin=c]{90}{Fine-Grained}}} 
        & \grayout{ImageNet Oracle} & \grayout{\texttimes}
        & & \grayout{67.2} & \grayout{62.2} & \grayout{68.6} & \grayout{62.9}
        & & \grayout{65.5} & \grayout{64.8} & \grayout{55.2} & \grayout{65.5}
        & & \grayout{60.0} & \grayout{60.7} & \grayout{58.3} & \grayout{60.0} \\
        & $\alpha$-ReQ &\grayout{\checkmark}
        & & 66.7 & 47.8 & \best{68.5} & 66.5
        & & 63.8 & 59.4 & 60.6 & 59.4
        & & 59.5 & 41.4 & 57.9 & 41.4 \\
        & RankMe &\grayout{\checkmark}
        & & \best{67.2} & \best{62.2} & 64.9 & 62.9
        & & 63.8 & \best{64.8} & \best{64.5} & \best{63.8}
        & & 59.5 & 56.9 & 57.9 & 59.5 \\
        & LiDAR & \grayout{\texttimes}
        & & 66.7 & \best{62.2} & 66.6 & \best{66.6}
        & & 68.3 & 62.4 & 64.3 & 62.4
        & & 58.3 & 60.7 & 58.2 & 58.3 
        \\
        & \cellcolor{skylightblue!35}\name{} & \cellcolor{skylightblue!35}\grayout{\checkmark}
        & \cellcolor{skylightblue!35} & \cellcolor{skylightblue!35}61.8 & \cellcolor{skylightblue!35}58.3 & \cellcolor{skylightblue!35}64.9 & \cellcolor{skylightblue!35}65.0
        & \cellcolor{skylightblue!35} & \cellcolor{skylightblue!35}\best{63.9} & \cellcolor{skylightblue!35}62.4 & \cellcolor{skylightblue!35}\best{64.5} & \cellcolor{skylightblue!35}62.4
        & \cellcolor{skylightblue!35} & \cellcolor{skylightblue!35}\best{60.0} & \cellcolor{skylightblue!35}\best{60.0} & \cellcolor{skylightblue!35}\best{58.3} & \cellcolor{skylightblue!35}\best{60.0} \\
        \bottomrule
    \end{tabular}
    }
    \vspace{0.5em}
    \caption{
        \textbf{Unsupervised model selection with \name{}.}
    We evaluate \name{} for hyperparameter selection against 
    a \grayout{supervised linear probe on ImageNet-1k}, two 
    unsupervised baselines: $\alpha$-ReQ~\cite{alpha2022Agrawal} and RankMe~\cite{garrido2023rankme},
    and a weakly-supervised one: LiDAR~\cite{thilaklidar}. 
    For each SSL objective (and architecture), hyperparameters are jointly 
    selected across according to \Cref{eq:select-hparam}.
    `Fine-grained' denotes the average performance across 
    all furhter datasets (i.e. iNat-21,SUN, Aircraft, CUB, CIFAR-10) 
    excluding ImageNet-1k. 
    \best{Bold} values indicate the top-performing model 
    selected by the criteria.
    \name{} is competitive with unsupervised and weakly-supervised baselines across SSL methods, including VICReg, 
    despite RankMe directly mirroring its objective.
    }
    \label{tab:hparam-selection}
\end{table*}

%% file: tables/compute.tex
\begin{table}
    \centering
    \resizebox{0.95\linewidth}{!}{%
    \begin{tabular}{@{} ll rr r rr @{}}
        \toprule
        \multirow{2}{*}{\textbf{Model}} & \multirow{2}{*}{\textbf{Architecture}} & \multirow{2}{*}{$D$} &\textbf{Param } & \textbf{Linear Probe} & \textbf{\name{}} \\
         &  & &\textbf{(M)} & \textbf{(min)} & \textbf{(min)} \\
        \midrule
        VICReg & ResNet-50 & 2048 & 24 & 64.5 & 2.1 & \\
        \addlinespace[0.3em]
        \multirow{4}{*}{DINOv2} & ViT-S & 384 & 22 & 65.8 & 1.8  \\
        & ViT-B & 768 & 86 & 64.5 & 2.1 &  \\
        & ViT-L & 1024 & 303 & 113.3 & 4.9 &  \\
        & ViT-G & 1408 & 1000 & 322.7 & 12.8 &  \\
        \bottomrule
    \end{tabular}%
    }
    \vspace{0.5em}
    \caption{\textbf{Computational cost.}
    Wall-clock time (min) on ImageNet-1k
    to evaluate frozen representations using either linear probing
    (10 epochs, batch size 1024) or \name{}
    (single feature-extraction pass followed by \dimmst{} computation).
    Rows vary the backbone architecture,
    primarily changing the output dimension $D$ and the number of parameters.
    Results are averages over 3 runs.}
    \label{tab:compute}
\end{table}

%% file: sections/05_conclusion.tex
\section{Conclusion}

In this work, we introduced \name{}, an unsupervised criterion 
for evaluating self-supervised representations based on 
the estimation of the intrinsic dimension via minimum spanning trees (\dimmst{}). 
Building on the theoretical connection between intrinsic dimension and generalization, 
we demonstrated that \name{} serves as a robust 
and an efficient geometric proxy for downstream performance in SSL. 

Our empirical evaluation across diverse datasets, 
architectures, and SSL objectives shows that \name{} 
correlates with supervised linear probing accuracy.
Furthermore, \name{} provides a principled label-free metric 
for hyperparameter selection, performing on par with supervised 
oracles and generalizing across heterogeneous SSL paradigms. 

By offering a unified framework for assessing SSL representations
without requiring annotated data, our work highlights the potential 
of intrinsic geometric descriptors to complement standard evaluation protocols. 

\textbf{Future Work.}
Future work could explore several directions. First, the relationship between intrinsic and effective dimensions (e.g., RankMe~\cite{garrido2023rankme}) warrants closer study.
As further discussed in \Cref{sec:effective-vs-intrinsic}, a large gap between the two
may signal curvature in the representation manifold, opening the door to
differential-geometric tools for complementary analysis.
Second, integrating intrinsic dimension estimates directly into SSL training
objectives could enable geometry-aware learning, guiding models toward
representations that are simultaneously compact and well-spread.
Finally, extending \name{} to dense tasks, e.g., segmentation
or generation, is a natural next step, as the underlying
\mst{} estimator is task-agnostic, though such settings do not
fall within the theoretical framing leveraged here.




%% file: sections/X0_dim_mst.tex
\section{\name{}'s Implementation Details}
\label{sec:supmat-idest-implementation}

\paragraph{Minimum Spanning Tree.}

\begin{figure}[htbp]
  \centering
  \includegraphics[width=0.7\linewidth]{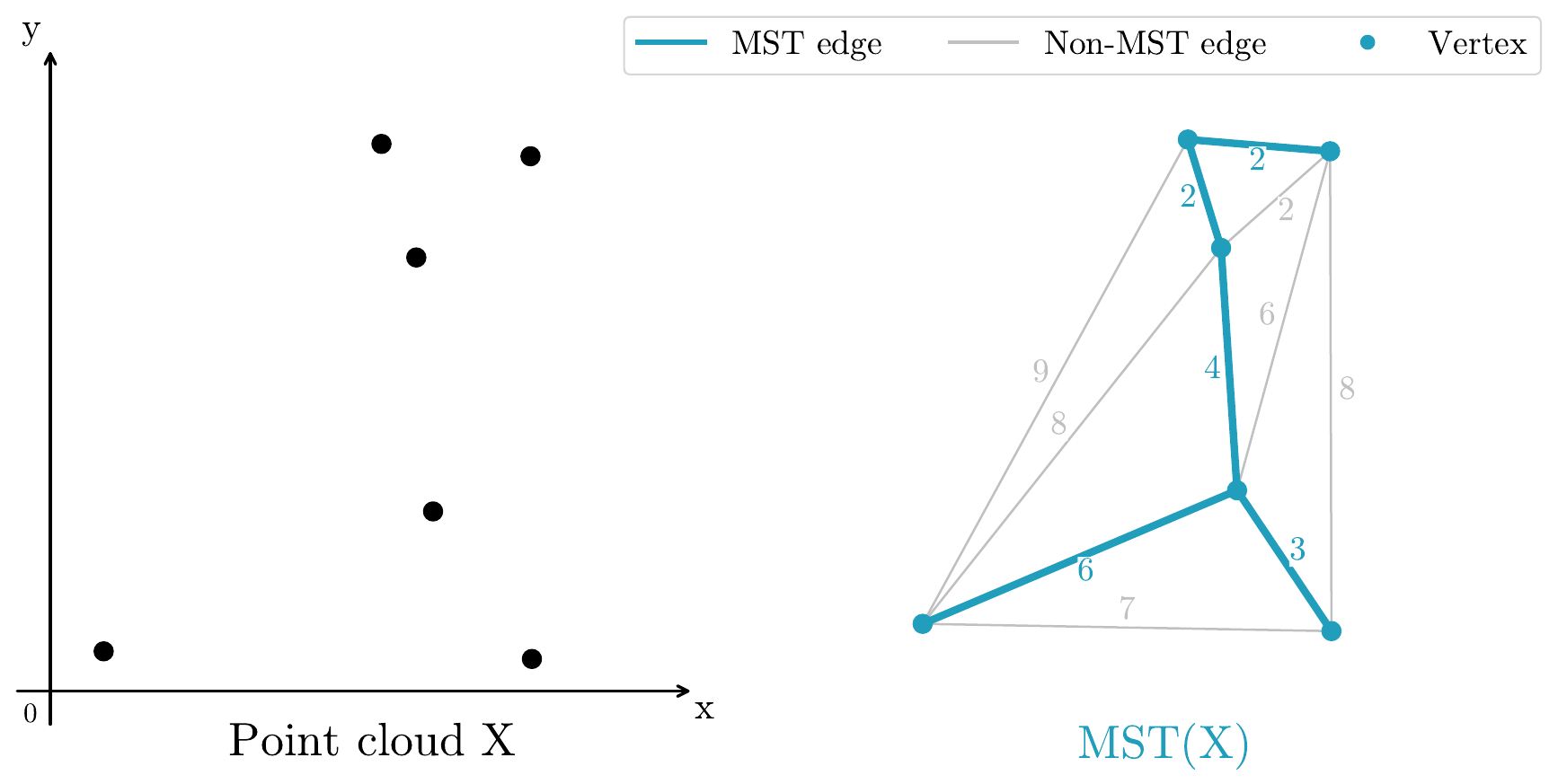}
  \caption{
       \textbf{Overview of a Minimum Spanning Tree (\mst).}
       \textit{(Left)} A point cloud $X \subset \mathbb{R}^2$. 
       \textit{(Right)} The \textcolor{bluegreen}{\mst{$(X)$}} connects all points 
       without forming any cycle, 
       while minimizing the total edge length.
       \textit{Grey edges} indicate pairwise connections not retained in the \mst{}: 
       including them would either create a cycle or increase the total length.
   }
   \label{fig:rw-mst}
\end{figure}

We complement the definition of the Minimum Spanning Tree (\mst{}) given
in~\Cref{subsection:id-estimators} with a visual overview in~\Cref{fig:rw-mst}
and with formal definitions below.

\begin{definition}{(Spanning Tree).}
  A spanning tree of $X$ is an undirected graph $G = (V, E)$ 
  with vertex set $V = X$ and edge set $E \subseteq V \times V$ 
  such that $G$ is connected and acyclic.
\end{definition}

\begin{definition}{(Minimum Spanning Tree).}
  A minimum spanning tree (\mst) of $X$ is a spanning tree 
  $G = (V, E)$ of minimum total edge weight:
  \begin{equation*}
      \costmst{X} \;\coloneqq\; \sum_{(u,v) \in E^*} \|u - v\|_2.
  \end{equation*}
\end{definition}

\paragraph{\dimmst{}.}
As described in~\Cref{subsection:id-estimators}, \name{} leverages \dimmst{} 
to estimate the intrinsic dimension of the representation space.
In practice, the ID is estimated by log-log linear regression over subsamples of 
increasing size: given subsamples $Z_{n_i}$ with sizes $n_i$, we fit
$\log(\costmst{Z_{n_i}}) \approx \frac{d-1}{d}\log(n_i) + \log(C)$,
and recover $d = 1/(1-m)$ from the fitted slope $m$.
The complete algorithm is given in~\Cref{alg:dimmst}.

\begin{algorithm}[H]
  \caption{Computation of \dimmst{}}
  \label{alg:dimmst}
  \begin{algorithmic}
    \STATE {\bfseries Input:} The set of representations $Z = \{z_1, \dots, z_N\}$, minimum sample size $n_{\min}$, skip size $\Delta$
    \STATE {\bfseries Output:} The estimated dimension: $\dimmst{}(Z)$
    \STATE Initialize $n \gets n_{\min}$, $E \gets []$
    \WHILE{$n < N$}
    \STATE $Z_n \gets \text{sample}(Z, n)$ \textcolor{bluegreen}{ \# random sampling of $n$ points}
    \STATE $E[i] \gets \costmst{Z_n}$ \textcolor{bluegreen}{ \# computation of the minimum spanning tree}
    \STATE $n \gets n + \Delta$
    \ENDWHILE
    \STATE $m,b \gets \text{linear regression}(\log(n_\text{min}: \Delta : N), \log(E))$ \textcolor{bluegreen}{ \# linear regression of the log-log plot}
   \STATE \dimmst{}(Z) $\gets 1 / (1 - m)$
    
  \end{algorithmic}
\end{algorithm}

We follow the implementation 
of~\citet{adamsFractalDimensionMeasures2020,dupuis2023generalization} for \dimmst{}.
For each dataset and model, we compute the dimension estimator.
Preprocessing (distance matrix computation and sorting) 
dominates in practice and is optimized via 
Ripser~\cite{tralie2018ripser,bauer2021ripser}, 
which implements an efficient sparse distance pipeline.
To keep computation tractable,
\name{} operates on a subsample of size $N \ll N_D$, 
where $N_D$ denotes the full dataset size; we set $N = 50{,}000$ throughout. 

%% file: sections/X2_pretraining.tex
\section{Implementation Details for SSL models studied}\label{sec:supmat-ssl-models}

\subsection{Image Classification Training Details}
\paragraph{Datasets.}
We evaluate the global quality of the SSL 
models using the widely adopted linear probing evaluation.
We consider:
\begin{enumerate}[itemsep=-0.3em, topsep=-0.1em]
    \item ImageNet dataset~\cite{deng2009imagenet}
    \item Large-scale fine-grained datasets: iNat-18~\cite{van2018inaturalist}, iNat-21~\cite{van2021benchmarking}, SUN397~\cite{xiao2010sun}
    \item Small-scale fine-grained datasets: CIFAR-10 and CIFAR-100~\cite{krizhevsky2009learning}, Aircraft~\cite{maji2013fine}, CUB200~\cite{welinder2010caltech}
\end{enumerate}

\paragraph{Evaluation protocol.}
For each baseline, we follow the protocol of~\cite{simeoni2025dinov3} 
and train a linear layer on the final frozen representation.
To obtain the frozen representation, we follow
each method's standard evaluation protocol, e.g., the \texttt{CLS} token 
after the layer norm, or \texttt{avgpool} if there is no \texttt{CLS} token.
Specifically, we use SGD with a momentum of $0.9$, and train 
for 10 epochs with a batch size of 1024, using random-resized-crop
data.
We perform the following grid search:
\begin{itemize}
    \item \textbf{Learning Rate} in $\{0.0001, 0.0002, 0.0005, 0.001, 
    0.002, 0.005, 0.01, 0.02, 0.05, 0.1, 0.2, 0.3, 0.5, 1\}$
\end{itemize}
 
For the fine-grained dataset (i.e., Aircraft), following~\cite{oquab2023dinov2},
we use a lighter weight evaluation using scikit-learn's LogisticRegression 
implementation with the L-BFGS solver.

\subsection{Models studied in~\Cref{subsection:predict}}
\input{tables/supmat-models.tex}

The checkpoints used were either downloaded from the original \href{https://github.com/}{gihub} or  \href{https://huggingface.co/timm}{timm}.
The complete list of pretraiend models is reported in~\Cref{tab:model-used}.

\subsection{Models trained in~\Cref{subsection:hparam}}
\label{subsec:hparam-models}
The complete list of models pretrained can be found below.
\paragraph{VICReg~\cite{bardes2022vicreg}.}
VICReg maximizes the informational content of 
embeddings by regularizing their empirical covariance matrix.

VICReg's loss is defined with 
three components:  \textit{(i)} a term 
to encourage the variance (diagonal of the covariance matrix) inside the current batch to be equal to 1, 
preventing collapse with all the inputs mapped on the same vector; 
\textit{(ii)} and a correlation regularization, encouraging the off-diagonal coefficients of the empirical covariance matrix to be close to 0, decorrelating the different dimensions of the embeddings. 
\textit{(iii)} an invariance loss that matches positive pairs 

We pre-trained ResNet-50 for 100 epochs using LARS, the projector used is an MLP with intermediate dimensions (8192, 8192, 2048), with a batch size of 2048, following the protocol of~\cite{garrido2023rankme}
\begin{enumerate}
    \item \texttt{lr}: $\text{wd} = 1e-6$, $\text{lr} \in \{0.1, 0.2, 0.3, 0.4,0.5\}$, $\text{inv}:25$, 
     $\text{cov}:5$, $\text{var}:25$
    \item \texttt{wd}: $\text{lr} = 0.3 $, $\text{wd} \in \{1e-7, 1e-6, 1e-5, 1e-4, 1e-3\}$, $\text{inv}:25$, 
     $\text{cov}:5$, $\text{var}:25$
     \item \texttt{cov.}: $\text{lr} = 0.3$, $\text{wd} = 1e-6 $, $\text{inv}:25$, $\text{cov}:5$, $\text{var}:25$ 
     $\text{cov}\in \{0.4, 0.6, 0.8, 1, 4, 16\}$, $\text{var}:25$
\end{enumerate}
    
\paragraph{DINO~\cite{caron2021emerging}.}
DINO uses a student-teacher framework.
Two versions of the same network (the student and the teacher) are fed different views of the same image. The student is trained to match the teacher's output probability distribution. To prevent the model from collapsing (i.e., giving the same output for every image), DINO uses a unique centering and sharpening operation on the teacher’s outputs. The teacher’s weights are updated as an exponential moving average of the student’s weights.

We pre-trained ViT-S for 100 epochs using Adam-W. The projector used is an MLP with intermediate dimensions (8192, 8192, 32768), with a batch size of 2048:
\begin{enumerate}
    \item \texttt{lr}: $\text{lr} \in \{1.25e-4, 2.25e-4, 0.0025, 0.002, 0.0075
     \}$, $\text{t-temp.}=0.04$, 
    $\text{s-temp} = 0.07$
    \item \texttt{s-temp}: $\text{lr}= 0.002 $, $\text{t-temp.}=0.04$, 
    $\text{s-temp} = \{ 0.07, 0.1, 0.2, 0.3, 0.4 \}$, 
    \item \texttt{t-temp}: $\text{lr}= 0.002 $, $\text{t-temp.}=\{0.01, 
    0.02, 0.03, 0.04, 0.05\}$, 
    $\text{s-temp} = 0.07$,
\end{enumerate}
\paragraph{I-JEPA~\cite{assran2023self}}
I-JEPA is a Joint-Predictive Architecture. 
It uses a \textbf{context block}
to predict the representations of several target blocks from the same image.
The context encoder is a Vision Transformer (ViT), which only
processes the visible context patches. The predictor is a smaller
ViT which takes the context encoder output and, 
conditioned on positional tokens, predicts the representations of a
\textbf{target block} at a specific location. 
The weights of the target encoder
are updated at each iteration via an exponential moving average of
the context encoder weights.

We pre-trained ViT-B for 300 epochs with the same protocol as in the
original papers with a batch size of 4096:
\begin{enumerate}
    \item \texttt{lr}: $\text{wd sch} = [0.04-0.4]$, $\text{lr} \in \{4e-5, 8e-5,9e-5, 1e-4, 1.25e-4, 2e-4, 3e-4\}$, $\text{target block size} = \{0.15,0.2\}$, $\text{context block size} = \{0.85,1.0\}$
    \item \texttt{Target Size Block} (the size of the target block): $\text{wd sch} = [0.04-0.4]$, $\text{lr} = 1.25e-5$, $\text{target block size} \in \{\{0.075,0.2\}, 
    \{0.1,0.2\},  \{0.125,0.2\},  \{0.2,0.25\}, \{0.2,0.25\}\}\}$, 
    $\text{context block size} = \{0.85,1.0\}$
    \item \texttt{Context Size} (the size of the context block): $\text{wd sch} = [0.04-0.4]$, $\text{lr} = 1.25e-5$, $\text{target block size} = \{0.15,0.2\}$, 
    $\text{context block size} \in \{\{0.4,1.0\}, \{0.5,1.0\}, 
    \{0.65,1.0\}, \{0.75,1.0\}, \{0.90,1.0\} \}$
    
\end{enumerate}
    

%% file: tables/supmat-models.tex
\begin{table}
\begin{center}
      \resizebox{0.99\linewidth}{!}{
        \begin{tabular}{l ccccc ccc ccc ccc r}
          \toprule
          & &  & & & \multicolumn{5}{c}{\textbf{Top-1 Accuracy (\%)}} & & \\
          &   \textbf{Method} & \textbf{Architecture} & Repo & & ImageNet & v2 & SUN397 & iNat-18 & iNat-21 &  \\
          \midrule
          & BarlowTwins~\cite{zbontar2021barlow} & ResNet-50 & \href{https://github.com/facebookresearch/barlowtwins}{github} & & 72.9 &  60.3 & 71.6 & 38.1 & 53.4 \\ 
          \hline
          &  \multirow{5}{*}{CLIP~\cite{radford2021learning}} & ResNet-50 & \href{https://huggingface.co/timm/resnet50_clip.openai}{timm} & & 69.6 & 52.8 & 76.5 & 13.7 & 46.4 \\
          & & ViT-B/16 & \href{https://huggingface.co/timm/vit_base_patch16_clip_224.openai}{timm} & & 80.1 & 68.8 & 82.2 & 51.1	&	63.4 \\
          & & ViT-B/32 & \href{https://huggingface.co/timm/vit_base_patch32_clip_224.openai}{timm} & & 76.1 & 63.5 & 80.6 & 41.9 & 54.6 \\
          & & ViT-L/14 & \href{https://huggingface.co/timm/vit_large_patch14_clip_224.openai}{timm} & & 84.3 & 75.1 & 84.8 & 59.8 & 72.5  \\ 
          \hline
          &  \multirow{2}{*}{DINO~\cite{caron2021emerging}} & ViT-S/16 & \href{https://github.com/facebookresearch/dino}{github} & & 76.0 &  64.4 & 72.5 & 48.4 & 60.9 \\
          & &  ViT-B/16 & \href{https://github.com/facebookresearch/dino}{github}  & & 77.8 & 66.7 & 74.8 & 49.9 & 63.4 \\
          \hline
          & \multirow{5}{*}{DINOv2~\cite{oquab2023dinov2}} & ViT-S/14 & \href{https://github.com/facebookresearch/dinov2}{github} & & 80.4 & 70.4 & 79.5 & 61.5 & 73.5 \\
          & & ViT-B/14 & \href{https://github.com/facebookresearch/dinov2}{github} & & 83.8 &  74.9 & 81.7 & 68.3 & 80.3 \\
          & & ViT-B/14-reg & \href{https://github.com/facebookresearch/dinov2}{github} & & 83.9 & 75.1 & 81.9 & 67.4 & 79.4 \\
          & & ViT-L/14 & \href{https://github.com/facebookresearch/dinov2}{github} & & 85.7 & 77.7 & 82.8 & 71.3 & 83.1 \\
          & & ViT-L/14-reg & \href{https://github.com/facebookresearch/dinov2}{github} & & 86.1 & 77.9 & 83.3 & 71.0 & 83.3 \\
          & & ViT-G/14 & \href{https://github.com/facebookresearch/dinov2}{github} & & 86.2 & 78.1 & 82.9 & 74.0 & 85.3 \\
          \hline
          & \multirow{4}{*}{DINOv3~\cite{simeoni2025dinov3}} & ViT-S/16 & \href{https://github.com/facebookresearch/dinov3/tree/main}{github} & & 79.3 & 68.7 & 79.1 & 61.8 & 69.4 \\
          & & ViT-B/14 & \href{https://github.com/facebookresearch/dinov3/tree/main}{github} & & 84.5 & 75.0 & 82.9 & 70.6 & 79.4 \\
          & & ViT-L/16 & \href{https://github.com/facebookresearch/dinov3/tree/main}{github} & & 87.2 & 79.5 & 84.6 & 76.1 & 84.2 \\
          & & ViT-7B & \href{https://github.com/facebookresearch/dinov3/tree/main}{github} & & 88.4 & 81.4 & 85.4 & 80.2 & 88.7 \\
          \hline
          & \multirow{3}{*}{EVA~\cite{sun2023eva}} & EVA01-g-14 & \href{https://huggingface.co/timm/eva_giant_patch14_clip_224.laion400m_s11b_b41k}{timm} & & 86.1 & 71.5 & 85.5 & 64.9 & 79.5 \\
          & & EVA02-b-16 & \href{https://huggingface.co/timm/eva02_base_patch16_clip_224.merged2b_s8b_b131k}{timm} & & 83.6 & 73.3 & 83.1 & 55.7 & 70.5\\
          & & EVA02-E-14-plus & \href{https://huggingface.co/timm/eva02_enormous_patch14_plus_clip_224.laion2b_s9b_b144k}{timm} & & 87.6 & 75.6 & 86.5 & 70.6 & 82.7 \\
          \hline
          & Franca~\cite{venkataramanan2025franca} & ViT-L/14 (In-21k) & \href{https://github.com/valeoai/Franca}{github} & & 84.2 & 75.4 & 80.7 & 73.5 & 70.6 \\
          \hline 
          & \multirow{3}{*}{iBoT~\cite{zhou2021ibot}} & ViT-S/16 & \href{https://github.com/bytedance/ibot}{github} & & 76.9 & 65.6 &73.6 & 48.1 & 60.6 \\
          & & ViT-B/16 & \href{https://github.com/bytedance/ibot}{github} & & 79.2 & 68.1 & 75.2 & 48.6 & 63.4 \\
          & & ViT-L/16 & \href{https://github.com/bytedance/ibot}{github} & & 80.6 & 69.9 & 76.4 & 51.5 & 66.2 \\
          \hline
          & I-JEPA~\cite{assran2023self} & ViT-G/16 & \href{https://github.com/facebookresearch/ijepa}{github} & & 75.8 & 63.8 & 74.9 & 45.9 & 58.7  \\
          \hline
          & \multirow{3}{*}{PE-Core~\cite{bolya2025perception}} & S16-336 & \href{https://huggingface.co/timm/PE-Core-S-16-384}{timm} & & 70.9 & 51.2 & 75.6 & 38.6 & 54.6 & \\
          & & B14-224 & \href{https://huggingface.co/facebook/PE-Core-B16-224}{timm} & & 83.2 & 72.7 &  83.8 &  60.8 &  74.9 \\
          & & L14-336 & \href{https://huggingface.co/facebook/PE-Core-L14-336}{timm} & & 87.8 & 78.7 & 87.0 & 74.1 & 84.1 \\
          \hline
          & \multirow{2}{*}{SigLIP~\cite{zhai2023sigmoid}} & ViT-B-16-SigLIP & \href{https://huggingface.co/timm/ViT-B-16-SigLIP}{timm} & & 82.5 & 68.9 & 82.7 & 53.3 & 69.6 \\
          & & ViT-L-16-SigLIP-256 & \href{https://huggingface.co/timm/ViT-L-16-SigLIP-256}{timm} & & 86.0 & 73.8 & 85.1 & 59.4 & 70.8 \\
          \hline
          & VICReg~\cite{bardes2022vicreg} & ResNet50 & \href{https://github.com/facebookresearch/vicreg}{github} & & 73.0  & 59.9 & 71.7 & 40.2 & 53.8 \\
          \bottomrule
        \end{tabular}
      }
    \end{center}
    \caption{\textbf{Foundation Models Studied.} 
    Overview of the pretrained models evaluated in this work, 
    spanning diverse SSL and vision-language objectives
    and architectures (ResNet, ViT). 
    Top-1 accuracy is reported on ImageNet, ImageNet-v2, 
    and additional fine-grained datasets SUN397, iNat-18, iNat-21. 
    Model weights are loaded from the official 
    repositories or \texttt{timm}.}
    \label{tab:model-used}
\end{table}

%% file: sections/XX_further_results.tex
\section{Additional hyperparameter selection results.}
To further validate \name{}'s performance on ResNet, 
we conducted additional experiments on DINO with a ResNet-50 
varying \texttt{s-temp}
(the student temperature) and the \texttt{t-temp} (teacher temperature), 
and with the \texttt{all} column in which methods 
must select from the full pool of hyperparameter configurations. 
The results are reported in~\Cref{tab:hparam-dino-resnet}
\input{tables/hparam-dino-resnet.tex}

%% file: tables/hparam-dino-resnet.tex
\begin{table}[htpb]
    \centering
    \renewcommand{\arraystretch}{1.2}
    \resizebox{0.4\linewidth}{!}{
    \begin{tabular}{lc ccc}
        \toprule
        & \multicolumn{4}{c}{\textbf{DINO} (ResNet-50)} \\
        \cmidrule(lr){2-5}
        Method & \rotatebox[origin=c]{90}{Unsup.} & \texttt{s-temp.} & \texttt{t-temp.} & \texttt{all} \\
        \midrule
        \grayout{ACC-1 Bounds} &
        & \grayout{[57.9, 67.5]} & \grayout{[63.0, 68.4]} & \grayout{[57.9, 68.4]} \\
        $\alpha$-ReQ & \grayout{\checkmark}
        & 61.9 & 63.0 & 63.0 \\
        RankMe & \grayout{\checkmark}
        & 61.9 & 67.3 & \best{67.3} \\
        LiDAR & \grayout{\texttimes}
        & 65.5 & \best{67.3} & \best{67.3} \\
        \cellcolor{skylightblue!40}\name{} & \cellcolor{skylightblue!40}\grayout{\checkmark}
        & \cellcolor{skylightblue!40}\best{67.5} & \cellcolor{skylightblue!40}\best{67.6} & \cellcolor{skylightblue!40}\best{67.6} \\
        \bottomrule
    \end{tabular}
    }
    \vspace{0.5em}
    \caption{
        \textbf{Unsupervised model selection with \name{} for DINO with a ResNet-50 backbone.}
    We evaluate \name{} for hyperparameter selection against 
    a \grayout{supervised linear probe on ImageNet-1k}, two 
    unsupervised baselines: $\alpha$-ReQ~\cite{alpha2022Agrawal} and RankMe~\cite{garrido2023rankme},
    and a weakly-supervised one: LiDAR~\cite{thilaklidar}. 
    Hyperparameters are jointly 
    selected across all hyperparameter axes according to \Cref{eq:select-hparam}.
    \best{Bold} values indicate the top-performing model 
    selected by the criteria.
    }
    \label{tab:hparam-dino-resnet}
\end{table}

%% file: sections/X3_compute-cost.tex
\section{Compute cost}
All trainings were performed on H100 GPUs.
The total computational cost of the project, including training baselines, experiments, and ablation studies, amounts to approximately 12,000 GPU-hours.

%% file: sections/X4_further_benchmark.tex
\section{Further study of Foundation Models}

\subsection{Additional properties studied in Joint-Embedding Self-Supervised Learning}

\begin{figure}[h]
    \centering
    \includegraphics[width=0.5\linewidth]{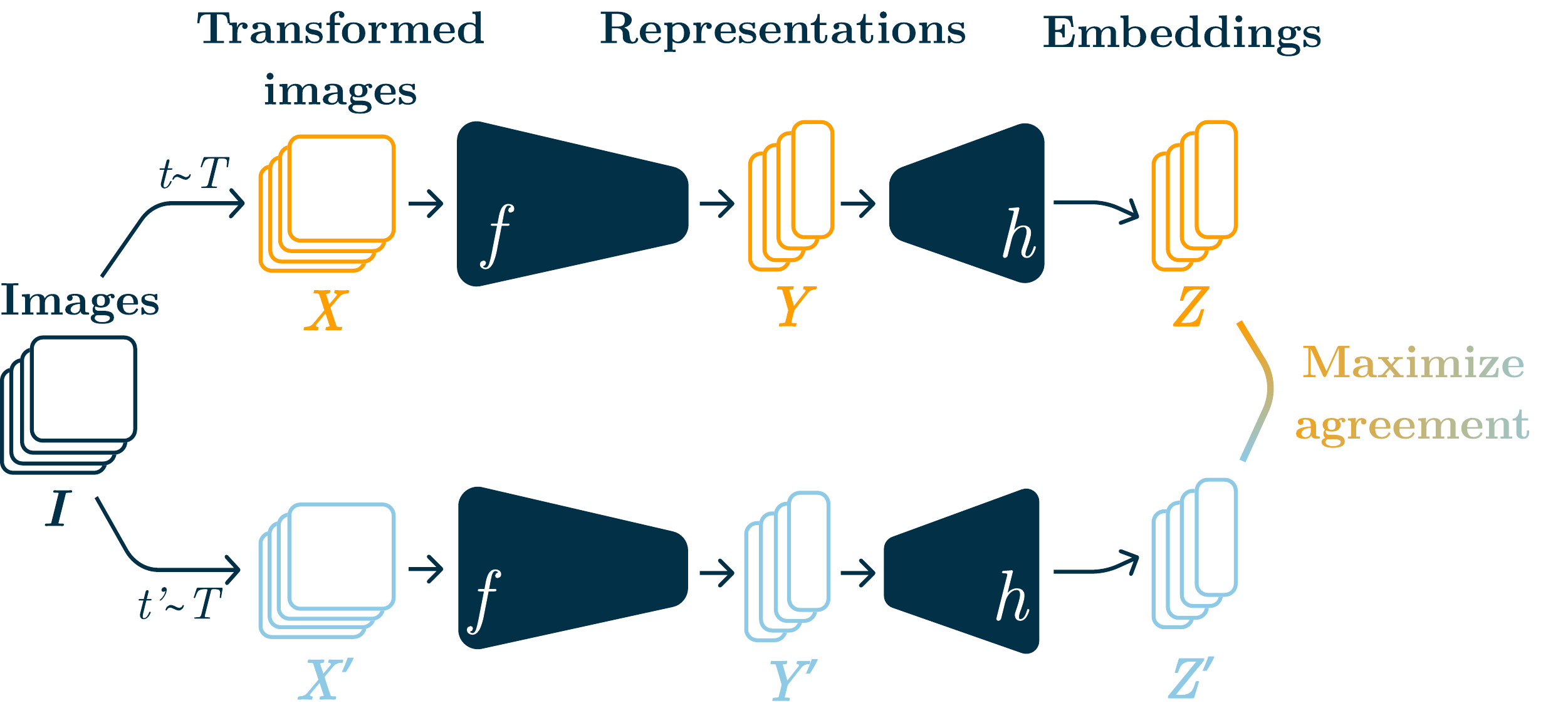}
    \caption{\textbf{Overview of Joint-Embedding Architectures.}
        Two separate data augmentation operators are sampled 
        from the same distribution 
        ($t, t' \sim \mathcal{T}$) and applied to each 
        image in a batch $I$ to obtain two views, $X$ and $X'$. 
        An encoder network $f$ and a projection head $g$ 
        are trained to maximize agreement between the resulting embeddings. 
        After training, the encoder $f$ and 
        its representations $Y$ are retained for downstream tasks.}
    \label{fig:je-ssl}
\end{figure}

A dominant approach in SSL  is \textit{joint embedding self-supervised learning} 
(JE-SSL)~\cite{chen2020simple,bardes2022vicreg}, 
where two networks are trained to produce similar embeddings for 
different views of the same image. An overview is presented in~\Cref{fig:je-ssl}.
Several metrics have been proposed to assess the quality of the learned representations in these methods.

\paragraph{Uniformity metrics.}
Uniformity is a desirable
properties of Joint-Embedding Self-Supervised 
methods~\cite{wang2020understanding,fang2024rethinking,mordacq2025tregs},
with the intuition that 
vectors should be roughly uniformly distributed on the unit hypersphere $S^{m-1}$,
preserving as much information of the data as possible.
Two main metrics have been proposed: 
\begin{itemize}
    \item \lu{}~\cite{wang2020understanding}, 
    based on the gaussian pairwise kernel: 
    \begin{equation}
        \lu = \log \underset{\substack{\text{i.i.d.} \\ x,y \sim p_\text{data}}}{\mathbb{E}}
        \exp^{-t || f(x) - f(y)||^2_2}, \, t > 0. 
    \end{equation}
    \item $\mathcal{W}_2$~\cite{fang2024rethinking}, the quadractic Wasserstein distance between the distribution of 
    the learned representation and $\mathcal{N}(0, I_M/m)$:
    \begin{equation}
        \mathcal{W}_2 := \sqrt{||\hat{\mu}||_2^2 + 1 
        + \text{tr}(\hat{\Sigma}) - \frac{2}{m}\text{tr}(\hat{\Sigma}^\frac{1}{2})
        }
    \end{equation}
    where $\hat{\mu}, \hat{\Sigma}$ are the sample mean and covariance mean. 
\end{itemize}
Though neither was designed with model comparison or 
hyperparameter selection in mind, we nonetheless 
investigate their potential across SSL paradigms.

\paragraph{RankMe~\cite{garrido2023rankme}.}
RankMe is formally the smooth rank measure, originally introduced
by~\citet{roy2007effective}:
\begin{equation}
    \text{RankMe}(Z) = \exp ( - \sum_{k=1}^{\min{(N,K)}} p_k \log p_k) \,, \text{with} \,
    p_k = \frac{\sigma_k(Z)}{||\sigma(Z)||_1} + \epsilon
\end{equation}
where $Z$ is the representations obtained.

\subsection{Foundation Models and further metrics.}

\begin{figure}
    \centering
    \begin{subfigure}{\linewidth}
        \centering
        \includegraphics[width=0.9\linewidth]{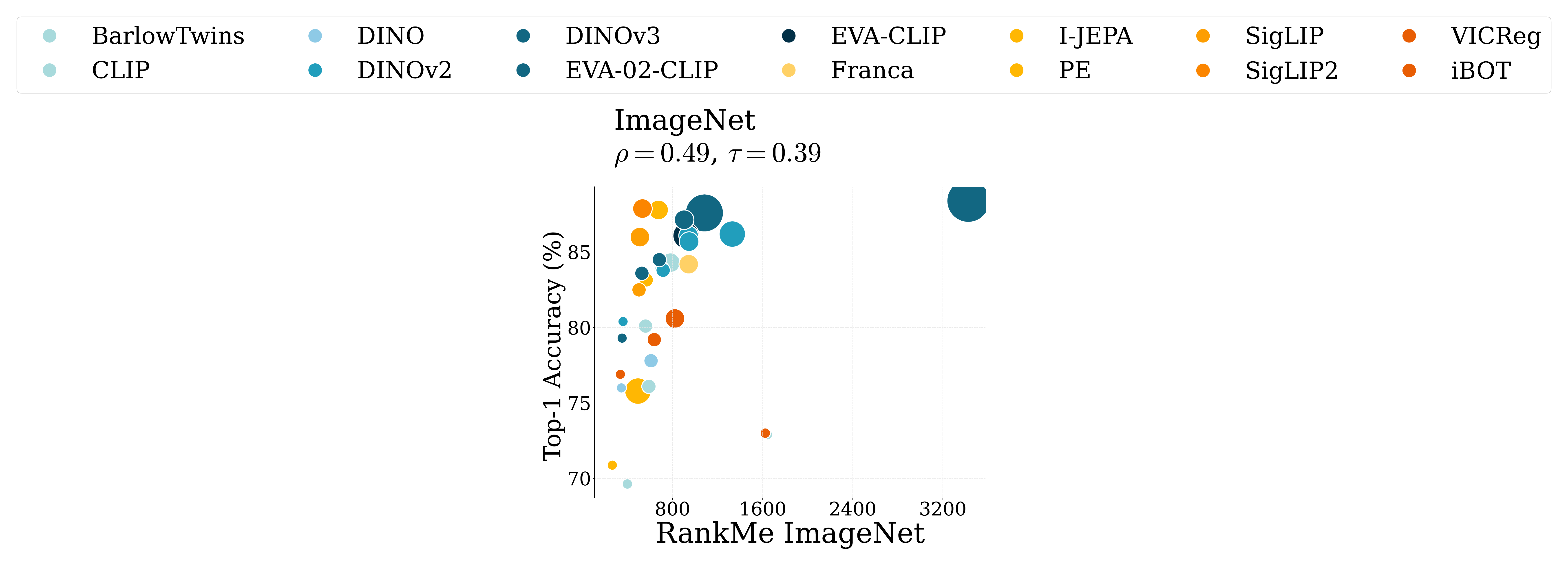}
        \caption{RankMe}
        \label{fig:supmat-rankme}
    \end{subfigure}
    \begin{subfigure}{\linewidth}
        \centering
        \includegraphics[width=0.9\linewidth]{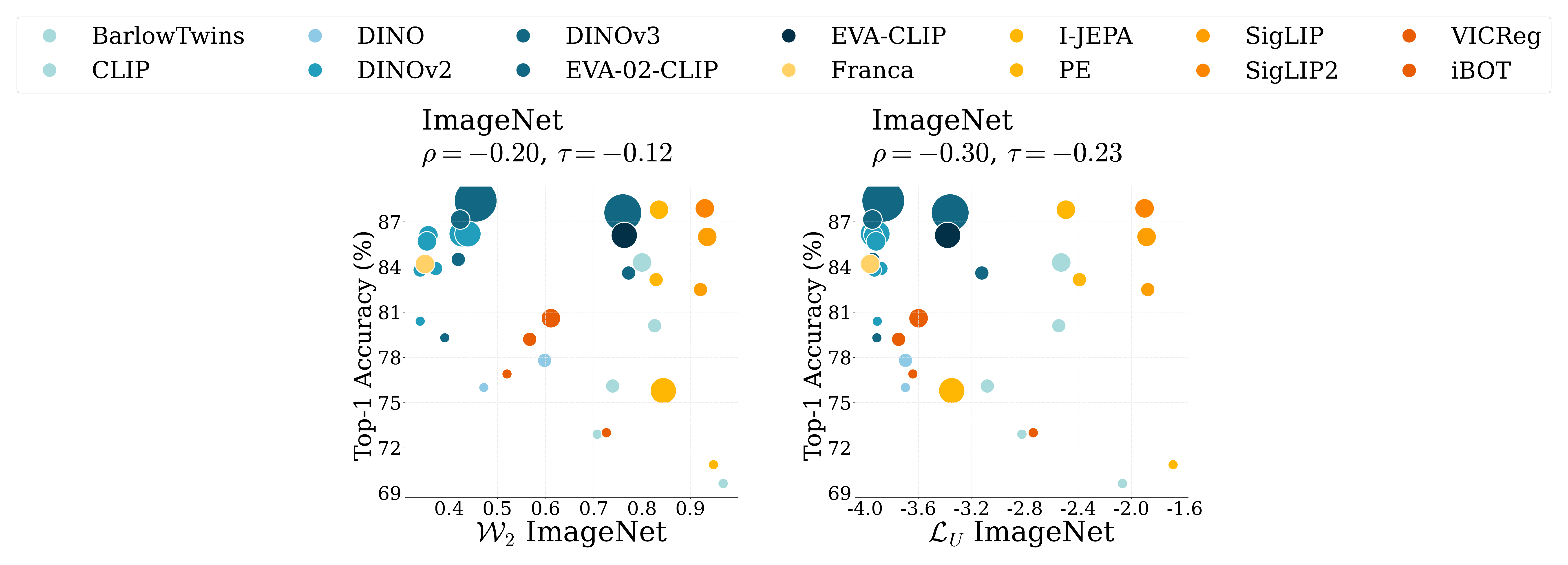}
        \caption{\lu and \wasser}
        \label{fig:supmat-unif}
    \end{subfigure}
    
    \caption{\textbf{Alternative metrics and Foundation Models.}
    Linear probing accuracy on ImageNet versus
    \textit{(Top)} RankMe and
    \textit{(Bottom)} uniformity metrics (\lu{} and \wasser{})
    for a diverse set of pretrained SSL models.
    }
    \label{fig:supmat-fms-futher-metrics}
\end{figure}

In~\Cref{fig:supmat-fms-futher-metrics} we study whether metrics
initially proposed for JE-SSL can
reflects representation quality across a diverse set of pretrained SSL models. 
Specifically, we compute RankMe, \lu, \wasser{} on frozen representations 
and compare it against standard linear probing accuracy on ImageNet.

For RankMe~\cite{garrido2023rankme},
the original study noted that it is not suited for comparisons across architectures.
A key reason is that RankMe is bounded by the output dimension of the model:
two models of different architectures that both span their full representation space
will yield different values, making cross-architecture comparisons unreliable (as highlighted
in~\Cref{fig:supmat-rankme}, where DINOv3 ViT-7B is a clear outlier compared
to DINOv3 ViT-L, despite similar ImageNet accuracies).

Concerning uniformity metrics (see~\Cref{fig:supmat-unif}),
neither \lu{} nor \wasser{} correlates consistently with linear probing accuracy across models.

Overall, all three metrics retain some predictive signal, as reflected by non-trivial
Kendall's $\tau$ and Spearman's $\rho$ values, yet their limited
correlation with accuracy motivates the search of alternative geometric proxies.

%% file: sections/X5_effective_intrinsic.tex
\section{Effective and Intrinsic Dimensions}
\label{sec:effective-vs-intrinsic}
\begin{figure}
    \centering
    \includegraphics[width=0.35\linewidth]{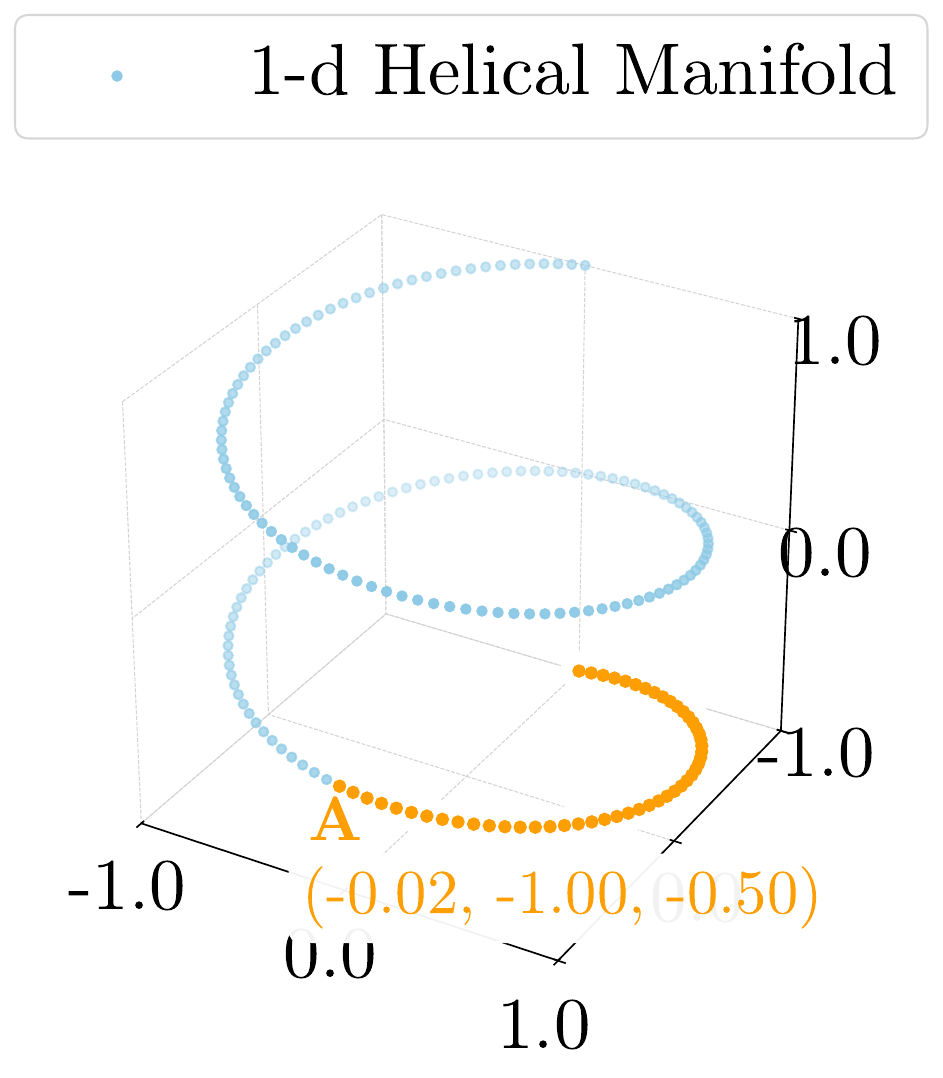}
    \caption{\textbf{Effective vs.\ intrinsic dimensionality.}
    A helix is embedded in a 3-dimensional ambient space (high effective dimension),
    yet is intrinsically a 1-dimensional manifold:
    point \textcolor{amberglow}{A} requires three coordinates to specify its position
    in the ambient space, but a single arc-length coordinate suffices to locate it
    along the curve.
    The two notions thus capture complementary aspects of geometry---how
    spread out a representation is versus how many degrees of freedom
    it truly contains.}
    \label{fig:supmat-effective-intrinsic}
\end{figure}

RankMe~\cite{garrido2023rankme} and \name{} capture complementary
notions of dimensionality.
The former measures \textit{effective rank}, the entropy of the singular-value
distribution of the feature matrix, reflecting how uniformly the representation
spreads across linear dimensions~\cite{roy2007effective}.
The latter estimates \textit{intrinsic dimension}: the minimum number of
degrees of freedom required to describe the data on its underlying manifold.
\Cref{fig:supmat-effective-intrinsic} illustrates this distinction with a helix,
whose points require three ambient coordinates yet lie on a curve
parameterised by a single value.

\begin{figure}
    \centering
    \includegraphics[width=\linewidth]{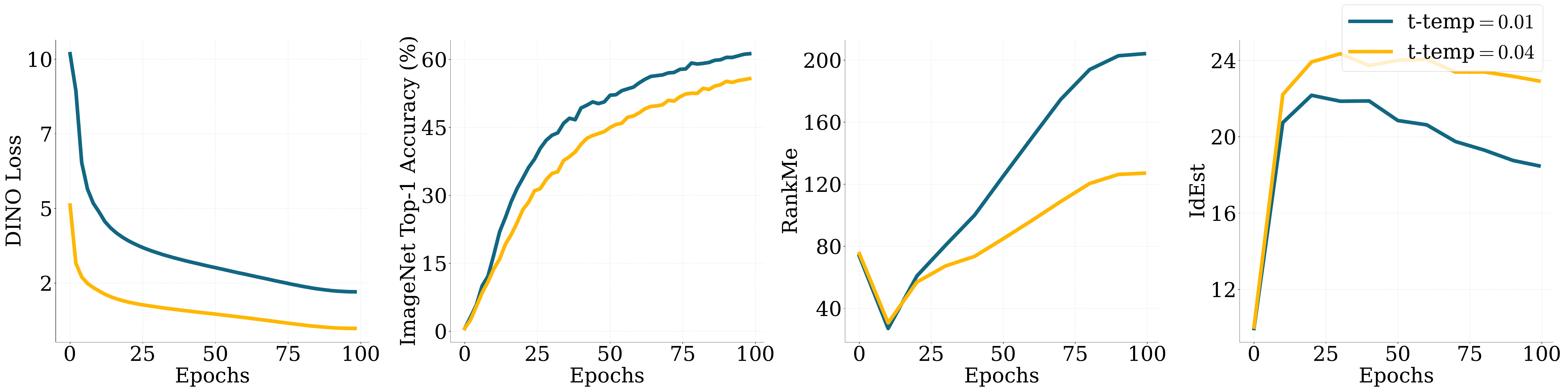}
    \caption{\textbf{Tracking Training Dynamics: RankMe and \name{}.}
    Evolution of the self-supervised loss, ImageNet-1k online classification top-1
    accuracy, \name{} and RankMe. 
    \name{} decreases while RankMe increases
    throughout training.
    }
    \label{fig:supmat-dino-training}
\end{figure}

RankMe correlates less strongly with downstream accuracy across foundation
models (\Cref{fig:supmat-fms-futher-metrics}; $\rho=0.49$, $\tau=0.39$)
than \name{} (\Cref{fig:fms-intra}, $\rho=-0.74$, $\tau=-0.55$),
confirming that intrinsic dimension captures information about representation
quality beyond what linear spread alone can reveal.
Yet RankMe's substantial correlation suggests that the effective rank of
representations remains a useful proxy for downstream performance.

\Cref{fig:supmat-dino-training} tracks the training dynamics of DINO (ViT-S)
and reveals a consistent trend: \name{} decreases while RankMe increases
throughout training.
This inverse relationship admits a natural information-bottleneck
interpretation: high-quality representations \emph{compress} the input onto a
compact, low-dimensional manifold (low intrinsic dimension) while
\emph{spreading} that information uniformly across ambient dimensions (high
effective rank) to avoid collapse.

As hypothesized in~\cite{ansuini2019intrinsic},
the gap between effective and intrinsic dimension relates to
the \emph{curvature} of the representation manifold: a flat manifold embedded
in $\mathbb{R}^d$ has matching intrinsic and effective dimensions, whereas a
highly curved one can occupy many ambient dimensions while remaining
intrinsically low-dimensional, as illustrated by the helix in
\Cref{fig:supmat-effective-intrinsic}.
Differential geometry offers a principled framework to formalise this gap;
curvature-aware metrics are a natural direction for future work to further
disentangle the structure of learned representations.